%% file: hybridGoalRecognition.tex
\documentclass[twoside,11pt]{article}
\usepackage{jair, theapa, rawfonts}

\usepackage{pgfplots}
\pgfplotsset{compat=newest}
\usetikzlibrary{matrix}

\usepackage{subfig}
\usepackage{algorithm}
\usepackage{algorithmic}
\usepackage{amsbsy}
\usepackage{amsmath}

\usepackage{booktabs}

\usepackage{url}

\DeclareMathOperator*{\argmax}{arg\,max}

\newtheorem{example}{Example}
\newtheorem{definition}{Definition}

\begin{document}

\title{Investigating the Combination of Planning-Based and Data-Driven Methods for Goal Recognition}

\author{\name Nils Wilken \email nils.wilken@uni-mannheim.de \\
        \addr Institute for Enterprise Systems, University of Mannheim \\ 69118 Mannheim, Germany
        \AND
       \name Lea Cohausz \email lea.cohausz@uni-mannheim.de \\
       \addr Data and Web Science Group, University of Mannheim \\ 69118 Mannheim, Germany
       \AND
       \name Johannes Schaum \email jschaum@mail.uni-mannheim.de \\
       \addr Institute for Enterprise Systems, University of Mannheim \\ 69118 Mannheim, Germany
       \AND
       \name Stefan Lüdtke \email stefan.lüdtke@uni-mannheim.de \\
       \addr Institute for Enterprise Systems, University of Mannheim \\ 69118 Mannheim, Germany
       \AND
       \name Heiner Stuckenschmidt \email heiner.stuckenschmidt@uni-mannheim.de \\
       \addr Data and Web Science Group, University of Mannheim \\69118 Mannheim, Germany
       }       

\maketitle
       
\begin{abstract}
An important feature of pervasive, intelligent assistance systems is the ability to dynamically adapt to the current needs of their users.
Hence, it is critical for such systems to be able to recognize those goals and needs based on observations of the user's actions and state of the environment.

In this work, we investigate the application of two state-of-the-art, planning-based plan recognition approaches in a real-world setting.
So far, these approaches were only evaluated in artificial settings in combination with agents that act perfectly rational.
We show that such approaches have difficulties when used to recognize the goals of human subjects, because human behaviour is typically not perfectly rational.
To overcome this issue, we propose an extension to the existing approaches through a classification-based method trained on observed behaviour data.
We empirically show that the proposed extension not only outperforms the purely planning-based- and purely data-driven goal recognition methods but is also able to recognize the correct goal more reliably, especially when only a small number of observations were seen.
This substantially improves the usefulness of hybrid goal recognition approaches for intelligent assistance systems, as recognizing a goal early opens much more possibilities for supportive reactions of the system.
\end{abstract}







\section{Introduction}
\label{sec:introduction}
The ultimate goal of smart assistance technologies is to dynamically adapt the infrastructure of a building to best meet the needs of their users by observing their behaviour and deducing their current needs.
Identifying users' goals and intentions based on their current and past activities is an important task in this context.
While there is some work on goal recognition in the context of smart assistance systems \cite{yordanova2017s,yordanova2019analysing,kruger2014computational}, so far research has mostly focused on recognizing users' current activities \cite{helaoui2013probabilistic,rashidi2010discovering,hoque2012aalo,yao2016learning,sztyler2017online}.
In this paper, we address the problem of identifying user goals as a basis for automatic support.
For this purpose, we look at the related problem of plan recognition, which is a long-standing topic in the Artificial Intelligence community \cite{kautz1986generalized,charniak1993bayesian}.
We believe that plan recognition methods are  particularly suited for this task as they do not only identify the goal a user intends to achieve, but also aim to recognize the most probable plan (i.e., ordered sequence of actions) for achieving this goal.
Knowing such a plan provides us with a better basis for supporting the user.

In this paper, we investigate the application of two state-of-the-art plan recognition approaches that are based on the principle of Plan Recognition As Planning (PRAP) \cite{ramirez2010probabilistic}.
More explicitly, the contributions of this paper are:
\begin{itemize}
    \item
    In Section \ref{sec:application}, we reveal and analyze some major shortcomings of PRAP approaches when applied to real-world scenarios.
    The main consequence of these shortcomings is that some goals can only be identified relatively closely before they are reached, which significantly reduces their potential benefits to an intelligent assistance system.
    \item
    As a possible solution to this problem, we propose a hybrid plan recognition method in Section \ref{sec:extension}.
    The proposed method combines the principle of PRAP with a data-driven probabilistic model that captures statistical relations between certain states of the environment and goals that can be learned from past observations.
    \item
    Finally, we empirically evaluate the proposed hybrid method in sections \ref{sec:evaluation} and \ref{sec:experimentalResults} and compare its performance to the performances of purely planning-based and purely data-driven approaches.
    The evaluation shows that both approaches can be applied to identify the goals of a user in real-world scenarios.
    Further, the results show that using a hybrid goal recognition method leads to a much earlier identification of the correct goal while only requiring very small amounts of training data. 
\end{itemize}


\section{Problem Definition} \label{sec:problem}
\textit{Probabilistic goal recognition} is the problem of inferring a probability distribution over a set of intended goals of an observed agent, given a possibly incomplete sequence of observed actions and a domain model that describes the domain in which the observed agent acts.
More formally, the aim of goal recognition approaches is to find a \textit{posterior} probability distribution $P(G|\mathbf{O})$ over all goals $g \in G$, given a sequence of observed actions $\mathbf{o}$.

\begin{figure*}[htbp]
    \centering
        \subfloat[]{\includegraphics[width=0.45\linewidth]{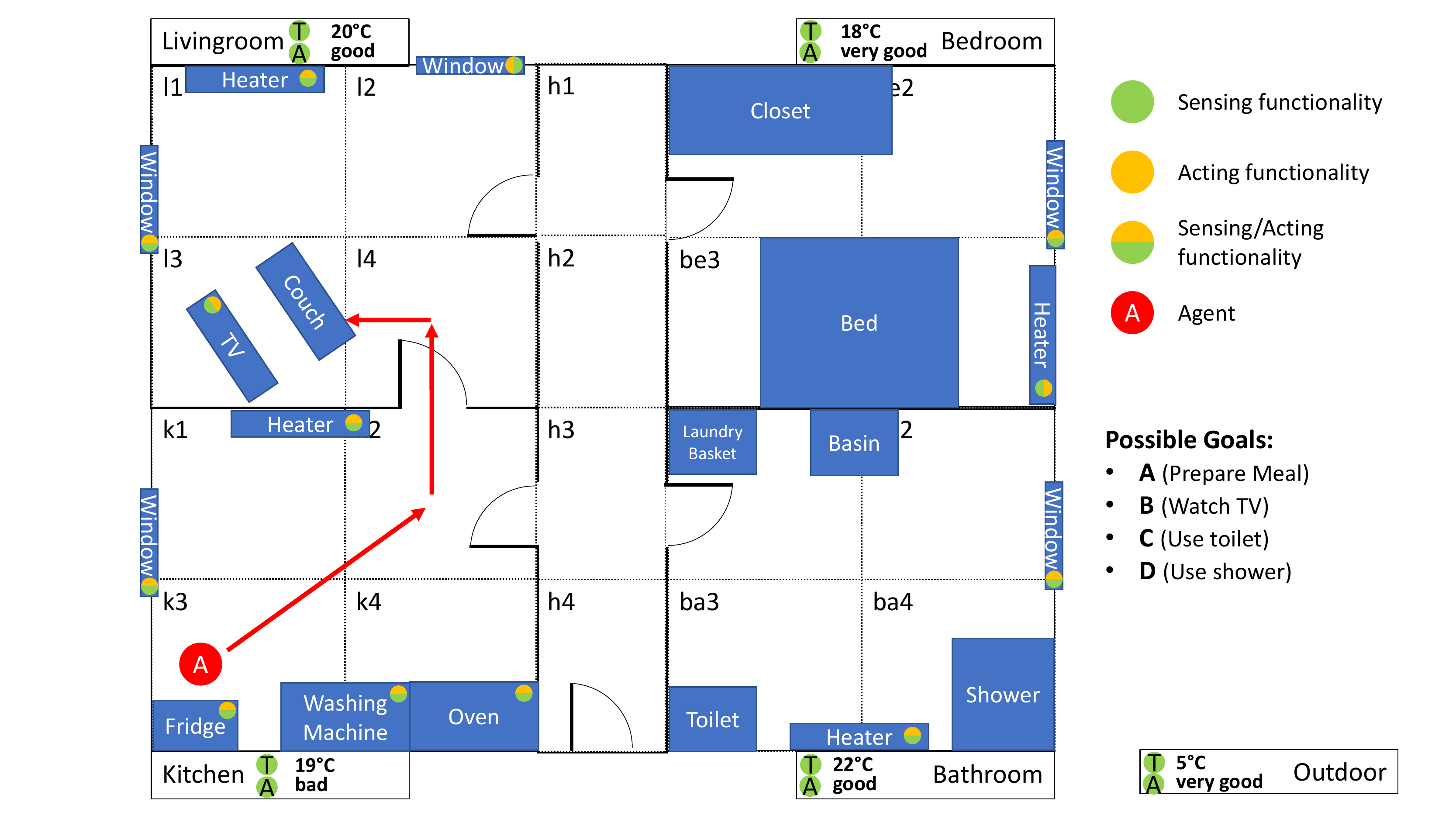}%
        \label{fig:scenarioA}}
    \hfil
        \subfloat[]{\includegraphics[width=0.45\linewidth]{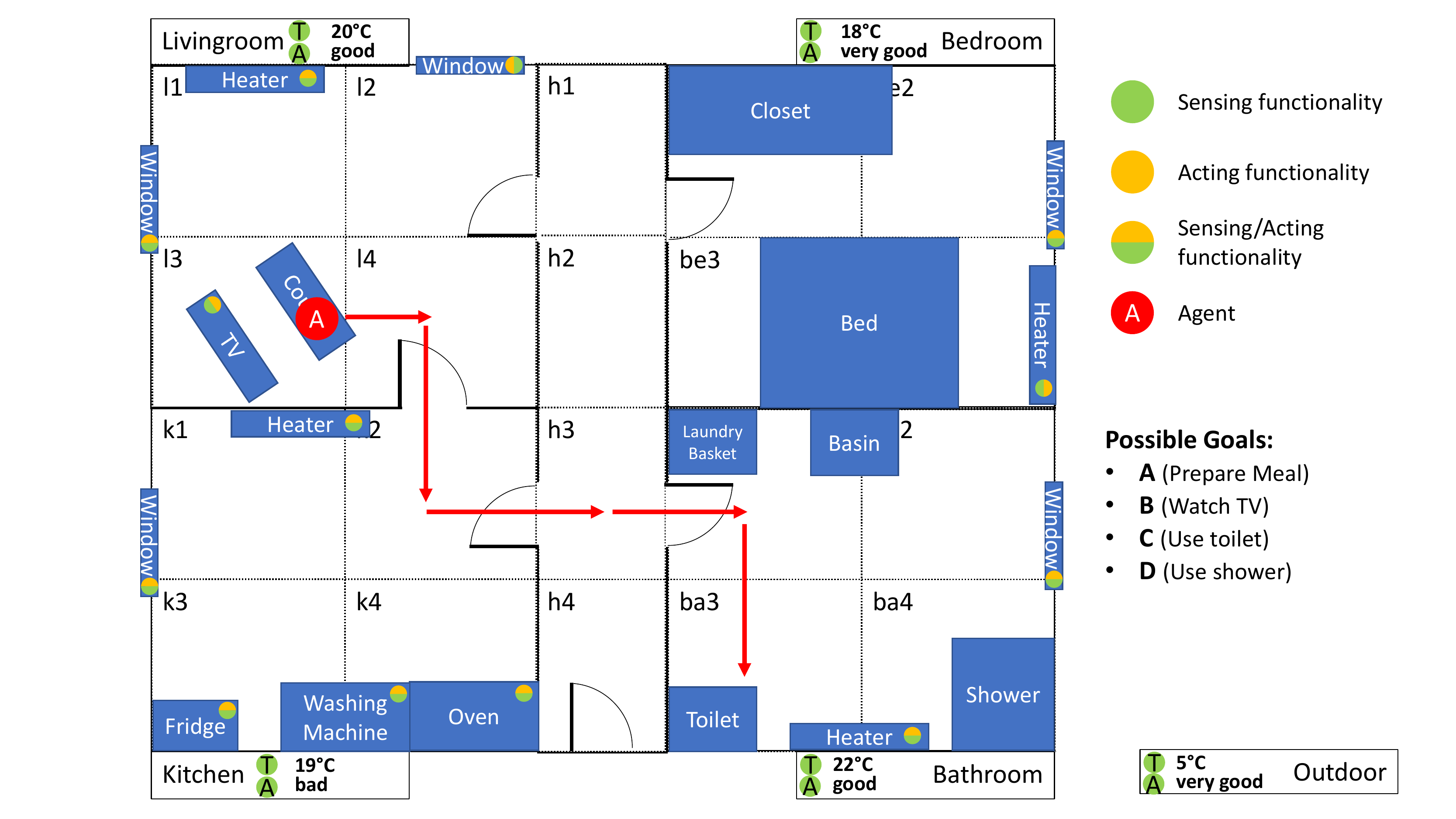}%
        \label{fig:scenarioB}}
    \caption{Illustration of an exemplary smart flat and a simple example use case (i.e., ``Beer Use Case'').}
    \label{fig:scenario}
\end{figure*}

This work considers the smart home domain as an example environment for goal recognition.
Figure \ref{fig:scenario} shows the layout of a smart flat and partial action sequences that sketch a simple use case.
This use case will be employed to analyze the shortcomings of the investigated planning-based goal recognition approaches.
It is important to note that this use case is completely synthetic and does not correspond to a real-world experimental setup.
The flat has four rooms and a hallway that connects all rooms.
In each room, different devices and furnishings are located.
Some of these objects can possibly function as sensor, actuator, or a mixture of both, which is indicated by the green and orange dots.
Furthermore, it is assumed that the current location of the agent can be sensed at all times and that the agent can navigate the cells in the flat by moving in all possible directions, including diagonal moves.

\begin{example}[Beer Use Case]
\label{exmpl:BUC}
Figures \ref{fig:scenarioA} and \ref{fig:scenarioB} roughly sketch parts of a small use case in this smart flat, which we will refer to as ``Beer Use Case'' (BUC) from here on.
In the BUC, a single agent is initially located in the cell ``l3'' in the livingroom, moves to the fridge, takes out a beer, and moves back to the couch in the livingroom (Fig. \ref{fig:scenarioA}).
When the agent arrives at the couch in the livingroom, she sits down on the couch, opens the beer, and drinks it while watching TV.
After a while, the agent decides to get another beer from the fridge. 
When the second beer is empty, the agent gets up from the couch, moves back to the kitchen, and subsequently via the hallway to the bathroom to use the toilet (Fig. \ref{fig:scenarioB}).
\end{example}


\section{Background} \label{sec:recognition}
In this work, we investigate the application of two state-of-the-art approaches to plan recognition to a real-world goal recognition scenario.
In contrast to probabilistic goal recognition, \textit{probabilistic plan recognition} not only describes the problem of inferring a probability distribution over a set of goals, but also the probability distribution over a set of possible plans that an agent might follow to reach it's intended goal.
From a solution to a plan recognition problem, the solution of the corresponding goal recognition problem can be derived by only considering the goals of the recognized plans.
Plan recognition is a long standing research area in the Artificial Intelligence community.
Recent plan recognition systems mostly rely on the Plan Recognition As Planning (PRAP) \cite{ramirez2009plan} principle and hence, utilize symbolic planning systems to solve plan- and goal recognition problems.

\subsection{Symbolic Planning}
Symbolic planning is based on a symbolic model of the planning domain that defines possible actions, their preconditions and effects on the environment.
Given a current state and goals in terms of partial state descriptions, planning methods aim to construct an optimal plan for reaching the goals from the current state consisting of a (possibly partial) order of actions to be executed.
We adopt the formalization of a planning problem from \cite{ramirez2010probabilistic}.

\begin{definition}[Planning Problem]
A Planning Problem is a tuple $P = \langle F, s_0, A, \mathcal{G} \rangle$ where $F$ is a set of fluents (boolean statements about properties of the modeled environment), $s_0 \subseteq F$ and $\mathcal{G} \subseteq F$ are the initial state and the goal description and $A$ is a set of actions with preconditions $Pre(a) \subseteq F$ and lists of fluents $Add(a) \subseteq F$ and $Del(a) \subseteq F$ that describe the effects of an action $a$ in terms of fluents that are added and deleted from the current state.
Actions have a non-negative cost $c(a)$.
A state is described by the subset of fluents which are currently considered to be true.
A goal state is a state $s$ with $s \supseteq \mathcal{G}$.
An action $a$ is applicable in a state $s$ if and only if $Pre(a) \subseteq s$.
Applying an action $a$ in a state $s$ leads to a new state $s' = (s \cup Add(a) \setminus Del(a))$.
A solution for a planning problem (i.e., a plan) is a sequence of applicable actions $\pi = a_1, \cdots a_n$ that transforms the initial state into a goal state.
The cost of the plan is defined as $c(\pi) = \sum \limits_i c(a_i)$.
A plan is optimal if the cost of the plan is minimal.  
\end{definition}

This basic model has been extended in different directions.
In this paper, we make use of two extensions.
One allows us to specify goals of form $\neg f$ that claim that a certain fluent $f$ is absent in the goal state.
The other enables the use of a conditional effect of form $p \rightarrow q$, where $p$ and $q$ are single fluents.
This means that when an action $x$ has such a conditional effect, fluent $q$ only becomes true after the execution of $x$ when $p$ was true before the execution \cite{ramirez2010probabilistic}.

\subsection{Plan Recognition As Planning: State-of-the-Art}
As already mentioned, many recent plan- and goal recognition approaches rely onto the principle of Plan Recognition as Planning (PRAP), which was first introduced by Ramírez and Geffner \cite{ramirez2009plan}.
All approaches that follow this principle have in common that they utilize concepts from the area of classical planning to compute probability distributions over a set of possible plans or goals, respectively.

\begin{definition}[Probabilistic Plan Recognition Problem]
\label{def:PlanRecognitionProblem}
A probabilistic plan recognition problem is a tuple T = $\langle D, G, O, P(G) \rangle$ where $D = \langle F, s_0, A, \emptyset \rangle$ is a planning domain, $G$ is a set of possible goals $g \subseteq F$, $\mathbf{o} = o_1, \cdots o_m$, where $o_i \in A$ is a sequence of actions that have been observed and $P(G)$ is the prior probability distribution over $G$.
A solution to the probabilistic plan recognition problem is the conditional probability of the goals given the observation sequence $\mathbf{o}$ (i.e., $P(G=g|\mathbf{O}=\mathbf{o}) \forall g \in G$).
\end{definition}

\paragraph*{Estimating Goal Probabilities}
Both plan recognition methods that are used in this work are based on the idea of using Bayes Rule to compute the posterior probabilities of the goals: 
\begin{equation}
    P(G|\mathbf{O}) = \alpha P(\mathbf{O}|G) P(G)
    \label{eq:goalProbabilityGivenObservation}
\end{equation}
It is assumed that the prior probabilities $P(G=g)$ of goals $g \in G$ are given in the problem definition.
Hence, the problem of probabilistic goal recognition boils down to the estimation of $P(\mathbf{O}|G)$.
Both investigated approaches utilize symbolic planning systems to estimate this probability.

The idea behind this is based on the assumption that agents act perfectly rational and hence, use strictly optimal plans (i.e, plans that minimize costs) to achieve their goals.
Furthermore, it is assumed that the probability of a goal to be the agent's actual goal can be estimated by relating the costs of an optimal plan that includes a given observation sequence $\mathbf{o}$ and an optimal plan that does not include $\mathbf{o}$, while reaching a given goal $g \in G$.
This can be done because an optimal plan that does not have to fulfill the requirement of including $\mathbf{o}$ is, according to the planning domain, a perfectly rational plan from the given initial state to a given goal $g$.
Hence, when the costs of an optimal plan that includes $\mathbf{o}$ are higher, this means that the agent is taking a detour compared to a perfectly rational plan.
More precisely, Ramírez and Geffner \cite{ramirez2010probabilistic} propose to calculate $P(\mathbf{o}|g)$ as follows: 
\begin{equation}
    P(\mathbf{o}|g) = \alpha^\prime \frac{exp\{-\beta \Delta(g)\}}{1 + exp\{-\beta \Delta(g)\}}
    \label{eq:observationProbabilityGivenGoal}
\end{equation}
Where $\alpha^\prime$ is a normalization factor and $\Delta(g) = c(\mathbf{o},g) - c(\overline{\mathbf{o}},g)$ is the cost difference between an optimal plan for $g$ that satisfies $\mathbf{o}$ and an optimal plan for $g$ that does not satisfy $\mathbf{o}$.
The costs $c(\mathbf{o},g)$ and $c(\overline{\mathbf{o}},g)$ can be computed out of the box using classical planning systems.

\paragraph{Translating a Plan Recognition Problem into Planning Problems}
The two state-of-the-art plan recognition methods used in this paper were proposed by Ramírez and Geffner \cite{ramirez2010probabilistic} (referred to as ``RG'' from here on) and Vered et al. \cite{vered2016mirroring} (referred to as ``GM'' (Goal Mirroring) from here on).
They mainly differ in the way they transform the original planning problem, which is necessary to ensure that the computed plans fulfill some necessary requirements.

To compute the probabilities $P(\mathbf{O}|G)$, the RG approach compiles a plan recognition problem $T = \langle P, G, \mathbf{o}, Prob \rangle$ into $2|G|$ planning problems.
For each goal $g \in G$ the two planning problems $P_o(g)$ and $P_{\overline{o}}(g)$ have to be compiled and solved.
Classical planning systems naturally cannot handle the requirement of satisfying a given sequence of observed actions in a computed plan.
To ensure that the computed solutions fulfill this requirement, the original planning domain $D$ has to be slightly modified.

\begin{definition}[Transformation of the Planning Domain (RG)]
For a given planning domain $D = \langle F,s_0,A \rangle$ and a given observation sequence $o$, the transformed domain is defined as $D^\prime = \langle F^\prime,I,A^\prime \rangle$ with $F^\prime = F \cup \{p_{o_i}| o_i \in (o_i)_0^n\}$, where $p_{o_i}$ is a new fluent and the actions $o \in A^\prime$ that are in $\mathbf{o}$ have an additional effect $p_{o_i}$ when $i=0$ and $p_{o_{i-1}} \rightarrow p_{o_i}$ has to hold otherwise.
\end{definition}

For this transformation it is assumed that no action appears twice in $\mathbf{o}$.
When this is the case, the corresponding actions are duplicated and renamed to ensure that the order of observed actions is unmodified in the resulting plans.
Now the costs $c(\mathbf{o},g)$ and $c(\overline{\mathbf{o}},g)$ can be calculated by solving the planning problems $P_o(g) = \langle F^\prime,s_0^\prime,A^\prime,g \cup \{p_{o_n}\} \rangle$ and $P_{\overline{o}(g)} = \langle F^\prime,s_0^\prime,A^\prime,g \cup \{\neg p_{o_n}\} \rangle$.

\paragraph*{Goal Mirroring}
The main difference between RG and GM is the domain translation procedure:
While RG adapts the actions in a given planning domain, GM uses a different initial state to generate plans that embed $\mathbf{o}$ each time a new observation is observed.
\begin{definition}[Transformation of Planning Problem (GM)]
For a given planning domain $D = \langle F,s_0,A \rangle$ and a given observation sequence $\mathbf{o}$, the transformed domain is defined as $D^\prime = \langle F^\prime,s_0^\prime,A^\prime \rangle$ with $F^\prime = F$, where $s_0^\prime = s_0[[\mathbf{o}]]$ and $s[[ \mathbf{o}]]$ returns as a result the planning state that is obtained when the action sequence $\mathbf{o}$ is applied to a planning state $s$.
\end{definition}
When this transformation is completed, analogously to the RG approach, GM calculates the costs $c(\mathbf{o},g)$ and $c(\overline{\mathbf{o}},g)$.
However, in contrast to RG, GM assumes for the calculation that an optimal plan from $s_0$ to a goal $g$ can be obtained by concatenating $\mathbf{o}$ with a plan for $g$ that starts at the adjusted initial state $s_0^\prime$.
From such a plan, again the costs $c(\mathbf{o},g)$ can be determined.
Furthermore, GM does not generate plans that strictly do not embed $\mathbf{o}$, but instead computes an optimal plan from $s_0$ to each goal and uses the costs of these plans analogously to the costs of plans that do not embed $\mathbf{o}$ as RG does (i.e., $c(\overline{\mathbf{o}},g)$).
Apart from this, GM uses the same heuristic as RG (i.e., Equation \ref{eq:goalProbabilityGivenObservation}) to compute goal probabilities $P(G|\mathbf{O})$ from these costs.

One major benefit of GM compared to RG is that it is expected to be much more time efficient in the case of online probabilistic goal recognition.
This becomes increasingly important with increasing complexity of the involved planning problems.
\begin{definition}[Online Probabilistic Goal Recognition]
\label{def:onlineGoalRecognition}
We define \textit{online probabilistic goal recognition} as a special variant of the \textit{probabilistic goal recognition} problem defined earlier.
In online goal recognition, we assume that the observation sequence $\mathbf{o}$ is revealed incrementally.
More explicitly, we introduce the notion of time $t \in \{0, \dots, T\}$, where $T = |\mathbf{o}|$.
For every value of $t$, one probabilistic goal recognition problem $R(t)$ can be induced as $R(t) = \langle D, G, \mathbf{o_t}, Prob \rangle$ where $D = \langle F, s_0, A, \emptyset \rangle$ and $\mathbf{o_t} = \{o_i | 0 \leq i \leq t, o_i \in \mathbf{o}\}$.
A solution to the online probabilistic goal recognition problem are the conditional probabilities $P_t(G=g|\mathbf{o_t}); \forall g \in G, t \in [0,T]$.
\end{definition}
Hence, in the case of online probabilistic goal recognition, GM solves, due to the different transformation procedure, only $|G||O| + |G|$ planning problems instead of $2|G||O|$ planning problems that RG solves. 


\section{Case Study: Goal Recognition in the Beer Use Case} \label{sec:application}
In this section we evaluate the performance of the RG and GM goal recognition approaches when applied to the synthetic \emph{BUC} example (see Section \ref{sec:problem}).
Furthermore, we demonstrate and discuss some major limitations of them.

\subsection{Experimental Setup} \label{subsec:ApplyingTheModel}
For the experiments, we modeled a planning domain $D_{BUC}$ in the \textit{Planning Domain Definition Language} (PDDL) \cite{mcdermott1998pddl}.
The goal set of the corresponding plan recognition problems (see Definition \ref{def:PlanRecognitionProblem}) is defined as $G_{BUC} = \{g_{prepare\_meal}, g_{watch\_TV}, g_{use\_shower}, g_{use\_toilet}\}$.
Further, following the approach of Ramírez and Geffner \cite{ramirez2010probabilistic}, we assume uniform prior probabilities $P_{BUC}(G)$ for all goals in $G_{BUC}$.

Based on this experimental setup, we conducted two experiments \textit{E1} and \textit{E2} with both recognition approaches, which, however, differ in the observation sequences that are used to compile the involved planning problems.
For experiment \textit{E1}, the actions in the utilized observation sequence represent the entire BUC (see Example \ref{exmpl:BUC}).
For experiment \textit{E2}, to evaluate how much the goal probability estimates depend on information gained from the observations of the agent getting and drinking beer, only the last six actions of the observation sequence used in \textit{E1} are used (i.e., in \textit{E2} the observations of the agent getting and drinking beer are not included in the observation sequences).
The remaining setups are similar for both experiments.
To solve the planning problems compiled from these setups, we used the LAMA \cite{richter2008landmarks} planner, which is a satisficing planner that can be used either in \textit{greedy} or \textit{anytime} mode.
When used in \textit{greedy} mode, the planner stops immediately when a plan is found, whereas in \textit{anytime} mode LAMA returns the best plan found in a given time window.
Here, we used LAMA in anytime mode.

\begin{table}
\caption{Evaluation results for the RG and GM goal recognition approaches when applied to \textit{E1} and \textit{E2} with the LAMA planner in \textit{anytime} mode. The results for both approaches are identical for both, \textit{E1} and \textit{E2}. Each row describes the probabilities $P(G|\mathbf{O})$ for all goals $G \in G_{BUC}$ for different lengths of $\mathbf{O}$ ($|\mathbf{O}|$). $g_1=g_{prepare\_meal}$,  $g_2=g_{watch\_TV}$, $g3=g_{use\_shower}$, $g4=g_{use\_toilet}$.}
\label{tab:ResultsPlanningExample}
\centering
    \subfloat[Results for \textit{E1}]{
        \begin{tabular}{ccccc}
            & \multicolumn{4}{c}{$P(G|\mathbf{O})$} \\ \cline{2-5}
                $|\mathbf{O}|$ & $g_1$ & $g_2$ & $g_3$ & $g_4 $ \\ \midrule 
            28 + 0 & \textbf{0.25} & \textbf{0.25} & \textbf{0.25} & \textbf{0.25} \\ 
            28 + 1 & \textbf{0.319} & 0.043 & \textbf{0.319} & \textbf{0.319} \\ 
            28 + 2 & \textbf{0.331} & 0.006 & \textbf{0.331} & \textbf{0.331} \\ 
            28 + 3 & 0.063 & 0.001 & \textbf{0.468} & \textbf{0.468} \\ 
            28 + 4 & 0.009 & 0.0 & \textbf{0.495} & \textbf{0.495} \\ 
            28 + 5 & 0.002 & 0.0 & 0.268 & \textbf{0.73} \\ 
            28 + 6 & 0.001 & 0.0 & 0.119 & \textbf{0.88} \\
            \bottomrule
            \end{tabular}
    }
    \hspace{.5cm}
    \subfloat[Results for \textit{E2}] {
        \begin{tabular}{ccccc}
            & \multicolumn{4}{c}{$P(G|\mathbf{O})$} \\ \cline{2-5}
            $|\mathbf{O}|$ & $g_{1}$ & $g_2$ & $g_3$ & $g_4 $ \\ \midrule 
            0 & \textbf{0.25} & \textbf{0.25} & \textbf{0.25} & \textbf{0.25} \\ 
            1 & \textbf{0.316} & 0.052 & \textbf{0.316} & \textbf{0.316} \\ 
            2 & \textbf{0.329} & 0.012 & \textbf{0.329} & \textbf{0.329} \\ 
            3 & 0.106 & 0.002 & \textbf{0.446} & \textbf{0.446} \\ 
            4 & 0.018 & 0.0 & \textbf{0.491} & \textbf{0.491} \\ 
            5 & 0.003 & 0.0 & 0.349 & \textbf{0.648} \\ 
            6 & 0.001 & 0.0 & 0.192 & \textbf{0.806} \\
            \bottomrule
        \end{tabular}
    }
\end{table}

\subsection{Results and Limitations} \label{subsec:PRAPShortcomings}
The results of the experiments are shown in Table \ref{tab:ResultsPlanningExample}.
Each row reports the probabilities $P(G|\mathbf{O})$ for all $g \in G_{BUC}$ for different lengths of $\mathbf{O}$.
As the RG and GM approaches are based on the same underlying principle and the \emph{BUC} domain is small enough to be handled efficiently by both approaches, the results for the experiments \textit{E1} and \textit{E2} are identical for both of them.
Hence, we only report one result table for each experiment.
The results of the experiments reveal two major limitations of the two planning-based approaches.

\begin{enumerate}
    \item In both experiments, the correct goal $g_4$ is only recognized at $|\mathbf{O}|= 28 + 5$ and $|\mathbf{O}|=5$ respectively, i.e., shortly before the goal is actually reached.
    This timestep corresponds to the observation where the agent has moved to the location \textit{ba3}, which is also where the toilet is located.
    This circumstance significantly reduces the possible usefulness for an intelligent assistance system, as the goal is recognized too late to provide support through  adaptations of the environment.
    
    \item The additional information that is included in the observation sequence used for \textit{E1}, has no impact on the estimated probabilities $P(G|\mathbf{O})$.
    Instead, the estimates are only based on information that are gained from the up to last six actions that are observed afterwards. 
    Intuitively, however, the additional information should have an impact onto the estimate, because there exists a causal relation between drinking beer and the probability that this agent pursues the goal to use a toilet afterwards.
    Thus, it should be possible to recognize the correct goal much earlier in the observation sequence.
\end{enumerate}
The main reason for these limitations of the two planning-based approaches is that the additional actions that are contained in the observation sequence used for \textit{E1} are not strictly necessary to fulfill the preconditions of any action that is required to reach one of the possible goals in an optimal way, i.e., drinking beer is not a necessary requirement to visit the bathroom.
As a consequence, these observations have the same impact on the estimate of $P(G|\mathbf{O})$ for all possible goals, although they might contain valuable information about the true probability of $P(G|\mathbf{O})$.


\section{A Hybrid Goal Recognition Approach} \label{sec:extension}
As shown in the previous section, a significant shortcoming of PRAP based approaches is that causal relations between goals and observations in the environment cannot be exploited in general.
This is due to the fact that the costs of plans are used as the only criterion to estimate the probabilities $P(\mathbf{O}|G)$.
To solve this problem, we propose to combine these PRAP approaches with a data-driven probabilistic causal model of agent goals and observations.

\subsection{A Statistical Causal Model of Observations}
\label{subsec:ACausalModelOfObservations}
We propose to model the relationship between goals and observations via a probability distribution $P(F_1, \dots, F_N | G)$, where $F_1,\dots,F_N$ are fluents of the planning state $s_t$.
This distribution models how goals (e.g., making a sandwich) affect the probability of specific fluents in the planning states (e.g., whether the agent holds a cucumber). 
In contrast to the PRAP models, the idea here is to \emph{learn} the parameters of $P(F_1, \dots, F_n|g)$ from training data. 
This way, the probabilistic model can capture the relations between fluents and goals that cannot be captured by planning-based approaches: Specifically, the planning-based models cannot capture statistical relations between fluents and goals that are not necessary for an optimal plan.
For example, in the BUC planning domain, drinking beer is not a necessary precondition for visiting the bathroom.
Still, drinking beer makes an eventual bathroom visit more likely. 

\begin{figure}[htbp]
    \centering
    \includegraphics[width=0.6\linewidth]{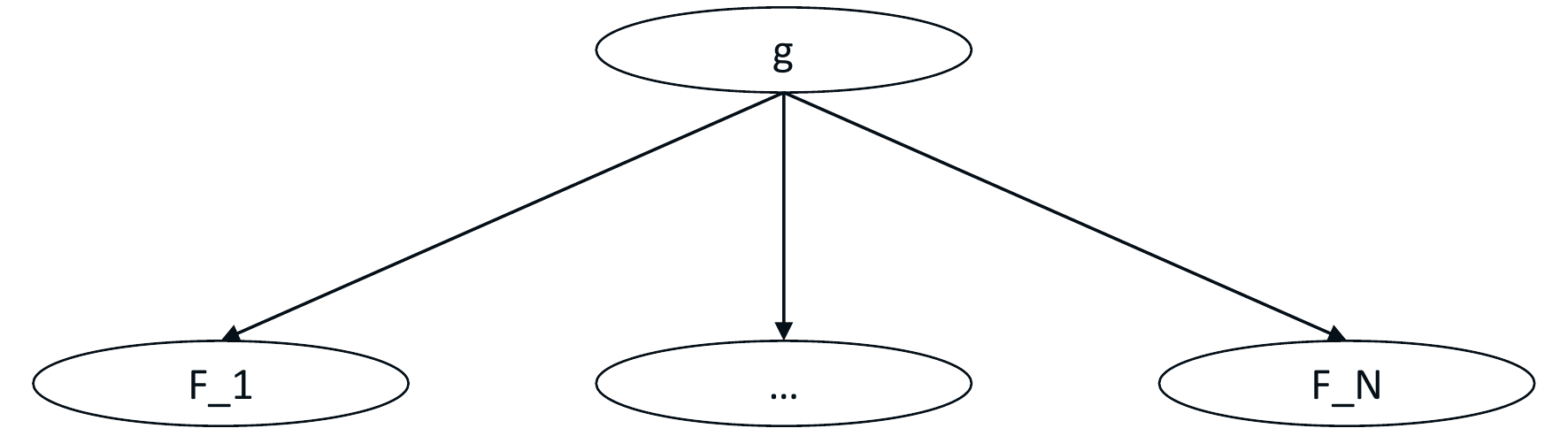}
    \caption{Bayesian network representation of the Naive Bayes Model.}
    \label{fig:BNModelTopology}
\end{figure}

In general, we could use any probabilistic model, like Bayesian Networks, deep generative models etc., to represent $P(F_1, \dots, F_N | G)$.
Here, we focus on a Naive Bayes model (NBM), i.e., assuming $P(F_1, \dots, F_n|g) = \prod_{i=1}^{n}{P(F_i|g)}$.
The model is visualized in Figure \ref{fig:BNModelTopology}. 
On the one hand, the strong independence assumptions between all fluents do not necessarily hold in practice. 
On the other hand, a Naive Bayes model has few parameters (linear in the number of variables).
Thus, training is possible even when training data is scarce -- as is often the case for activity sequences of real human subjects. 

Note that we made another simplifying assumption here: The distribution $P(F_1,\dots,F_N | G)$ only models the dependency of the \emph{current planning state} on the goal $g$, but not the dependency of the \emph{action sequence} $\mathbf{o}$ on $g$.
More specifically, different observation sequences that result in the same planning state could be associated with different goals, but we deliberately neglected this information to make the learning problem tractable.

\subsection{A Hybrid Goal Recognition Method}
\label{subsec:hybridGoalRecognitionMethod}
To overcome the shortcomings of the PRAP approaches discussed in Section \ref{subsec:PRAPShortcomings}, we propose to combine it with the proposed probabilistic learned model of $P(\mathbf{o}|g)$.

For the combination of the goal probabilities, we use model stacking, which is a common ensemble learning method.
In model stacking, a so-called meta-model is used to generate a combined estimate from the estimates of heterogeneous base models.
In this work, we studied two different, manually designed, meta-models to combine the estimate of one of the PRAP approaches and the estimate of the NBM.

It is important to note that there are some major differences between the ways in which the two utilized base models estimate the goal probabilities $P(G|\mathbf{O})$.
One of these differences is that the PRAP models are based purely on manually specified knowledge (in the form of the planning domain), whereas the NBM only relies on this manually defined domain knowledge to establish its structure but \emph{learns} its parameters from training data.
Another major difference between the models is the set of features which are used to estimate the probability $P(\mathbf{O}|G)$ and hence, also the probability $P(G|\mathbf{O})$.
Although the predictions of both models are based on an observed action sequence and an observed initial state of the environment, they estimate the probability $P(\mathbf{O}|G)$ based on different features that can be derived from these observations.
The planning-based methods use two entire plans to estimate the probability $P(\mathbf{o}|g)$.
In contrast, the NBM model uses only the planning state that results from applying the sequence $\mathbf{o}$ to the initial state to estimate $P(\mathbf{o}|g)$. 

\paragraph*{Weighted Sum of Predictions}
As a first meta-model, we use a \textit{weighted sum} (WS) of the two base-model estimates from one of the planning-based approaches and the NBM.
More formally, this meta-model is defined as follows:
\begin{equation}
    \label{eq:weightedSum}
    P(g|\mathbf{o}) = w_sP_s(g|\mathbf{o}) + w_dP_d(g|\mathbf{o})
\end{equation}
$P_s$ is the goal probability estimate of one of the symbolic PRAP approaches, $w_s$ is the weight for the symbolic approach, $P_d$ is the goal probability estimate of the data-driven NBM, and $w_d$ is the weight of the data-driven NBM.

\paragraph*{Tiebreaking of the PRAP approaches}
The second meta-model, which we refer to as \textit{tiebreaking} (TB), only considers the estimate of the NBM when more than one goal is assigned with the highest likelihood by the PRAP approach.
The intuition behind this meta-model is based on the observations from the results of experiments \textit{E1} and \textit{E2}: Here, the PRAP approaches never ranked wrong goals as most probable, but not \textit{only} ranked the true goal as most probable.
When this is the case, we combine the two estimated probabilities of the NBM and the planning-based approaches again by taking the weighted sum of them.
More formally, the meta-model is defined as follows:
\begin{equation}
P(g|\mathbf{o}) = \\
\begin{cases}
P_{s}(g|\mathbf{o}),&\text{if } |\argmax_{g \in \mathcal{G}} P_{s}(g|\mathbf{o})| = 1\\
w_sP_s(g|\mathbf{o}) + w_dP_d(g|\mathbf{o}),&\text{if } |\argmax_{g \in \mathcal{G}} P_{s}(g|\mathbf{o})| > 1,
\end{cases}
\end{equation}

\paragraph*{Hybrid Goal Recognition Performance for E1}
To evaluate whether the proposed hybrid method is able to leverage on the additional information contained in the observation sequence used for \textit{E1}, we applied the hybrid approach to \textit{E1}.
For the experiments, we assumed both weights $w_s$ and $w_d$ to be 0.5 and, due to the lack of training data for the \emph{BUC} use case, modeled the parameters of the data-driven model manually.
The results of the experiments are summarized in Table \ref{tab:ResultsCausalExp1}.
\begin{table}
\caption{Evaluation results for \textit{E1} for both meta-models. Each row describes the probabilities $P(\mathbf{O}|G)$ for all goals $G \in G_{BUC}$ for different lengths of $\mathbf{O}$ ($|\mathbf{O}|$). $g_1 = g_{prepare\_meal}$,  $g_2=g_{watch\_TV}$, $g_3=g_{use\_shower}$, $g_4=g_{use\_toilet}$.}
\label{tab:ResultsCausalExp1}
\centering
    \subfloat[Results for the \textit{tiebreaking} (TB) meta-model.]{
    \label{tab:ResultsCausalExp1Tiebreaking}
        \begin{tabular}{ccccc}
            & \multicolumn{4}{c}{$P(G|\mathbf{O})$} \\ \cline{2-5}
            $|\mathbf{O}|$ & $g_{1}$ & $g_{2}$ & $g_{3}$ & $g_{4} $ \\ \midrule 
            28 + 0  & 0.131 & \textbf{0.61} & 0.13 & 0.13 \\
            28 + 1  & 0.287 & 0.176 & 0.214 & \textbf{0.323} \\
            28 + 2  & 0.293 & 0.158 & 0.22 & \textbf{0.33} \\
            28 + 3  & 0.146 & 0.14 & 0.283 & \textbf{0.431} \\
            28 + 4  & 0.005 & 0.16 & 0.361 & \textbf{0.474} \\
            28 + 5  & 0.002 & 0.0 & 0.268 & \textbf{0.73} \\
            28 + 6  & 0.0 & 0.0 & 0.12 & \textbf{0.88} \\
            \bottomrule
        \end{tabular}
    }
    \hspace{.5cm}
    \subfloat[Results for the \textit{weighted sum} (WS) meta-model.] {
        \label{tab:ResultsCausalExp1Sum}
        \begin{tabular}{ccccc}
            & \multicolumn{4}{c}{$P(G|\mathbf{O})$} \\ \cline{2-5}
            $|\mathbf{O}|$ & $g_{1}$ & $g_2$ & $g_3$ & $g_4 $ \\ \midrule
            28 + 0 & 0.131 & \textbf{0.61} & 0.13 & 0.13 \\ 
            28 + 1 & 0.287 & 0.176 & 0.214 & \textbf{0.323} \\ 
            28 + 2 & 0.293 & 0.158 & 0.22 & \textbf{0.329} \\ 
            28 + 3 & 0.146 & 0.14 & 0.283 & \textbf{0.431} \\ 
            28 + 4 & 0.005 & 0.16 & 0.361 & \textbf{0.474} \\ 
            28 + 5 & 0.037 & 0.088 & 0.165 & \textbf{0.71} \\ 
            28 + 6 & 0.037 & 0.088 & 0.091 & \textbf{0.785} \\
            \bottomrule
        \end{tabular}
    }
\end{table}
The results show that goal $g_{4}$ is ranked as most probable once one additional observation is made for both meta-models.
Before this point, $g_{2}$ is considered to be the most probable goal.
Hence, the results show that a hybrid goal recognition model is able to leverage on the information that is contained in the observations that the agent drank beer.
Consequently, it can recognize the true goal much earlier than the PRAP approaches for this example.
Also interesting to note is that the results are identical for both meta-models until the fifth observation is observed.
This makes sense as the results in Table \ref{tab:ResultsPlanningExample} show that the PRAP approaches are undecided for the first four observation steps.
Hence, both meta-models use the same weighted sum to combine goal probability estimates.
For the fifth and sixth observation steps, both meta-models estimate the highest probability for the correct goal (i.e., $g_4$).
However, the TB meta-model estimates slightly higher probabilities for this goal and hence, is slightly more confident in $g_4$ being the actual goal of the agent.


\section{Evaluation Setup} \label{sec:evaluation}
This section describes the experimental setup of the empirical evaluation.
The empirical evaluation aims to achieve the following goals:
\begin{enumerate}
    \item Evaluate the performance of the planning-based methods, the NBM, and other well-known data-driven techniques, when applied to goal recognition problems in a real-world scenario, to determine which methods are best suited to be used in a hybrid goal recognition method.
    \item Show that a hybrid probabilistic goal recognition method is able to achieve superior performance, compared to purely planning-based and purely data-driven methods.
    \item Investigate how the performance of the hybrid method is affected by increasing the number of possible goals.
\end{enumerate}
We used real-world and artificial datasets for empirically evaluating the goal recognition methods.
In all experiments, the \emph{online} goal recognition problem was considered (as defined by Definition \ref{def:onlineGoalRecognition}).

\subsection{Datasets}
This section presents the real-world and artificial datasets that were used for evaluation. 

\paragraph*{Real-World Kitchen Dataset}
As a real-world data set, we used the CMU-MMAC Kitchen Dataset \cite{torre2009data} to which we will refer to as ``CMU'' from here on.
It contains data from different sources (e.g., video, motion capture, etc.) that were recorded by observing different people while cooking one out of five recipes.
We will consider the different dishes the observed participants might cook as possible goals.
We first transformed the existing ``raw'' data into a suitable format for our purpose.
As a starting point of this transformation, we used the results of a semantic annotation project \cite{yordanova2018data}.
In this project, planning domains in PDDL format and annotated observation sequences were created for three of the five recipes (i.e., brownies, eggs, and sandwich).
In addition, we created annotations for the remaining two recipes (i.e., pizza and salad).
Table \ref{tab:observationStatistics} displays some summarizing statistics of the observation sequence lengths per goal.
Note that the CMU dataset only includes the first five goals.
Interesting to note is that the average- and median observation sequence lengths substantially differ between the recipes.
In addition, the standard deviations of the sequence lengths are relatively high.
This indicates that the different observed persons used significantly different paths to reach one of the goals.

\begin{table}[h!]
\centering
\caption{Statistics of the observation sequence lengths per recipe in the CMU and artificially extended CMU datasets ($g_1$=brownies, $g_2$=eggs, $g_3$=sandwich, $g_4$=salad, $g_5$=pizza, $g_6$=bread, $g_7$=briocheBraid, $g_8$=cheeseburger, $g_9$=spaghetti, $g_{10}$=spinachFetaPastry). The original CMU dataset only contains the first five goals.}
\label{tab:observationStatistics}
\begin{tabular}{ccccccccccc}
          & $g_1$ & $g_2$  & $g_3$ & $g_4$  & $g_5$ & $g_6$ & $g_7$ & $g_8$ & $g_9$ & $g_{10}$ \\ \midrule
Average   & 111.17 & 87.08 & 57.85 & 110.56 & 90.40 & 69.73 & 50.27 & 122.96 & 67.63 & 108.12 \\
Median    & 108.0  & 88.0  & 58.0  & 108.5  & 89.5  & 70.0  & 50.0  & 114.0 & 65.5  & 96.5   \\
Std. Dev. & 15.43  & 17.25 & 9.02  & 17.94  & 15.35 & 7.57  & 7.61  & 27.29 & 9.43  & 39.01  \\
\bottomrule
\end{tabular}
\end{table}

One limitation of the CMU Dataset is that although it is based on sensor recordings of real human participants that were recorded while they were cooking different recipes, the general setup that was used during the sensor recordings is still rather artificial and, therefore, does not necessarily reflect all aspects of natural behaviour in a cooking scenario.
Nevertheless, we still think that it is able to provide a solid basis to judge whether the investigated goal recognition approaches are able to handle recognition scenarios of real-world complexity, which is the aim of this work.

\paragraph*{Extending the CMU Dataset with Artificially Generated Data} \label{subsec:dataSampling}
To evaluate the scalability of the proposed hybrid approach to a higher number of goals, we extended the data from the CMU dataset with five artificially generated goals.
In the remainder of this work, we will refer to this extended dataset as ``ACMU''.
The added goals (bread, brioche braid, cheeseburger, spaghetti, and spinach feta pastry) were manually defined in the planning domain.
This domain, which was used in both experiments (just with different sets of possible goals), contains 3411 different actions and 1627 different fluents.
To generate artificial observation sequences for these goals, which is required for training the data-driven methods, we developed a procedure to sample such observation sequences for a given planning problem.
The intuition underlying our proposed hybrid approach is the assumption that human behavior includes carrying out actions that are not strictly necessary in order to reach the current goal (i.e., which are not rational according to the planning domain).
Thus, the sampling procedure should reflect this intuition.
Consequently, it is not sufficient to use optimal plans as generated by existing planning systems: These plans generally only contain actions that are strictly necessary to reach a given goal.
Additionally, a planning system will always generate very similar plans when it is presented with the same planning problem.
As a solution, we propose a sampling algorithm that generates artificial observations sequentially.
At each step, the algorithm either returns the action used in an optimal plan, or a randomly drawn action (where the probability of drawing a certain action depends on the goal).
More details on the sampling algorithm can be found in Appendix \ref{sec:appendixSampling}.

Table \ref{tab:observationStatistics} presents some summarizing statistics for the artificially generated observation sequences for $g_6$ - $g_{10}$.
The average- and median lengths of the artificially generated observation sequences are comparable to the sequences that are based on real observed sensor data.
In addition, there is also a comparable amount of variation among these sequences, which is important to reflect the properties of real-world observations.

\paragraph*{Artificial Dataset}
To further investigate the scalability of the proposed hybrid approach and make the results better comparable to existing work, we used a well-known artificial planning domain, which was commonly used as a benchmark in the existing literature \cite{ramirez2010probabilistic},\cite{pereira2020landmark}.
This domain models a logistics problem, where different objects have to be delivered to several destinations.
In contrast to the CMU domain, this domain is much smaller and contains only 356 actions and 84 fluents.
We will refer to the resulting dataset as ``LOG'' from here on.
As this domain is purely synthetic, no real observation sequences existed.
Hence, we used the same sampling procedure that was used to extend the CMU dataset to generate artificial observation sequences for this domain.
Table \ref{tab:observationStatisticsLog} displays some summarizing statistics for the resulting observation sequences.
An important difference to the CMU datasets is that the average observations sequence length is much smaller in the logistics dataset, which is mainly caused by the synthetic nature of this domain.
Nevertheless, although this is the case, there is still a recognizable amount of variance among the sampled observation sequences.
\begin{table}[h!]
\centering
\caption{Statistics of the observation sequence lengths per recipe in logistics dataset.}
\label{tab:observationStatisticsLog}
\begin{tabular}{ccccccccccc}
          & $g_1$ & $g_2$  & $g_3$ & $g_4$  & $g_5$ & $g_6$ & $g_7$ & $g_8$ & $g_9$ & $g_{10}$ \\ \midrule
Average   & 21.33 & 22.47  & 23.23 & 23.43  & 21.40 & 22.80 & 23.67 & 22.27 & 24.27 & 22.37 \\
Median    & 21.0  & 21.0   & 23.0  & 22.0   & 21.0  & 22.0  & 23.0  & 22.0  & 23.5  & 22.0  \\
Std. Dev. & 2.31  & 3.51   & 2.64  & 3.48   & 2.93  & 2.51  & 2.27  & 2.80  & 2.32  & 4.52 \\
\bottomrule
\end{tabular}
\end{table}

\subsection{Goal Recognition Methods}
\label{subsec:goal-rec-exp-design}
In the empirical experiments, we applied different symbolic, data-driven, and hybrid goal recognition methods.

\paragraph*{Symbolic Goal Recognition Methods}
We used two state-of-the-art planning-based methods: GM and RG, which were presented in Section \ref{sec:application}.
To solve the planning problems that are generated by the two PRAP approaches, we used the MetricFF \cite{hoffmann2003metric} planner.
MetricFF is a satisficing planner that supports metric fluents.
Following Ramírez and Geffner \cite{ramirez2010probabilistic} and Vered et al. \cite{vered2016mirroring}, we assumed equal cost for all actions and used a value of $\beta=1$.
As the timeout for the MetricFF planner, we used 340 seconds.
After this timeout, the planner will be forced to stop and the associated planning problem is considered not solvable.

\paragraph*{Data-Driven Goal Recognition Methods}
We evaluated four different data-driven methods: The NBM presented earlier in Section \ref{sec:extension} and three additional data-driven approaches, which were selected following a recent study by Borrajo et al. \cite{borrajo2020goal}.
Specifically, we used K-Nearest-Neighbors (KNN) \cite[pp. 738-740]{russel2016modern}, XGBoost \cite{chen2016xgboost}, and a Long-Short-Term Memory (LSTM) network \cite{hochreiter1997long}.

For experiments with the KNN and XGBoost approaches, we evaluated two different data encodings.
First, we used a binary encoding of the planning states, consisting of the state of all planning fluents. 
Second, following the work of Borrajo et al. \cite{borrajo2020goal}, we used a vector encoding of the observed action sequence.

We performed a grid search to determine the hyper-parameters for the KNN and XGBoost methods.
For the planning fluent-based encoding, we used a leaf size of 30 and $k=8$ for KNN and a learning rate of 0.24, minimum child weight of 5, $\alpha=0.42$, $\lambda=1.15$, maximum depth of 5, and $\gamma=0.99$ for XGBoost.
In contrast, for the action vector encoding, we used a leaf size of 40 and $k=3$ for KNN and a learning rate of 0.31, minimum child weight of 2, $\alpha=3.26$, $\lambda=1.03$, maximum depth of 3, and $\gamma=0.96$ for XGBoost.

For the LSTM, we also evaluated two different data encodings: A one hot representation of both the observed action sequence and the observed state sequence.
Here, we used different vector encodings than for the other data-driven methods because LSTM models, in contrast to the other considered methods, are explicitly designed to handle temporal data sequences which are better captured by one hot data encodings \cite{borrajo2020goal}.
For both setups, we used the ADAM optimizer with a learning rate of 0.01, a batch size of 32, and 100 epochs.

\paragraph*{Hybrid Goal Recognition Methods}
We evaluated the two different meta-models described in Subsection \ref{subsec:hybridGoalRecognitionMethod}.
For both meta-models, we computed the weight for the NBM as \(w_{NBM}(n, t) = \frac{a}{1 + e^{-b(t - ((cn)+d))}}\), where $n$ is the number of training examples used to train the NBM, $t$ is the number of observations used for goal recognition, and $a$, $b$, $c$, and $d$ are fitting parameters.
As for the data-driven approaches, we performed a grid search to determine the best performing values for $a$, $b$, $c$, and $d$.
Accordingly, we set the parameters to $a=0.5$, $b=-0.15$, $c=4$, $d=2.5$ (CMU), and $a=0.5$, $b=-0.15$, $c=5$, $d=1$ (ACMU) respectively for the CMU datasets.
For the LOG dataset, we used the following parameter values: $a=0.2$, $b=-0.15$, $c=0$, and $d=0$.
The weight for the PRAP approaches is calculated as \(w_{PRAP} = 1 - w_{NBM}(n)\) for all datasets.

\subsection{Experimental Design}
\label{subsec:performanceMeasure}
To assess the goal recognition performance of the different methods, we used the mean goal recognition accuracy.
To calculate the accuracies, in contrast to most previous works, we did not consider a goal to be recognized correctly if it is part of a set of goals that were assigned with the highest likelihood.
Instead, we only considered a goal to be recognized correctly if it is the \emph{only} goal that was assigned with the highest probability.
We decided for this evaluation method as it, in our opinion, better reflects the usefulness of the prediction for practical application in an assistance system.
If such a system is provided with more than one most probable goal, it has to randomly decide for one goal.
Furthermore, as we consider online goal recognition problems in this evaluation, we calculated the mean accuracy for different fractions of total observations that were used for goal recognition.
Here we used relative numbers because the lengths of the involved observation sequences substantially differ.
Hence, the mean accuracy $Acc$ for a relative number of observations $\lambda \in [0, 1]$ is calculated as follows:
\begin{equation}
    Acc(\lambda,\mathcal{D}) = \frac{\sum_{R \in \mathcal{D}}{[R(\lfloor T_{R}\lambda\rfloor) = \Tilde{g_R}]}}{|\mathcal{D}|}
\end{equation}
Here, $\mathcal{D}$ is a set of online goal recognition problems $R$, $\Tilde{g_R}$ denotes the correct goal of goal recognition problem $R$, $T_R$ is the maximum value of $t$ for online goal recognition problem $R$ (i.e., length of observation sequence that is associated with $R$), and $[R(t) = \Tilde{g_R}]$ equals 1 if the correct goal is recognized for $R(t)$ and 0 otherwise.

To evaluate the performance of the symbolic goal recognition methods, we calculated $Acc(\lambda,\mathcal{D})$ for different values of $\lambda$ and for different domains $\mathcal{D}$.
To investigate the performance of the data-driven approaches in relation to the number of available training examples (i.e., number $n$ of complete observation sequences), we used a slightly adjusted cross-validation procedure:
For a given value of $n$, we split a set of online goal recognition problems $\mathcal{D}$ into $k$ partitions, where $k = |\mathcal{D}|/n$.
Then, $k$ models were trained, but in contrast to the typical cross-validation procedure, only \textit{one} of the partitions was used as the training set and the remaining partitions were used for validation.
In cases where $\mathcal{D}$ cannot be splitted into $k$ partitions of equal size $n$, we randomly sampled sequences from the other partitions to complete the partitions with a size smaller than $n$.
To assess the performance of a data-driven method, we calculated $Acc(\lambda,\mathcal{D})$ for all $k$ models and, subsequently, took the average over these accuracies.
For the evaluation of the hybrid methods, we calculated the combined estimates for all results obtained from the cross-validation procedure for the data-driven approaches and then also took the average of the accuracies of all $k$ models.


\section{Experimental Results} \label{sec:experimentalResults}
In the following, we present and discuss the results of the experiments corresponding to each of the three evaluation goals defined in Section \ref{sec:evaluation}.

\subsection{Symbolic and Data-Driven Goal Recognition}
\paragraph*{Symbolic Goal Recognition Results}
\input{figures/figures_tex/evaluationFigure_RG_GM_comparison}
We start by comparing the two planning-based goal recognition methods on the CMU dataset.
Figure \ref{fig:CMURGGMComparison} shows the average goal recognition accuracy of the RG and GM approaches on this dataset.
It can be seen that the GM approach outperformed the RG approach consistently, except for the case when only very small fractions of the observation sequences are used.
Furthermore, the accuracy of the RG approach decreases when more than 20\% of the observations are used.
The main reason for this behavior is the fact that the involved planning problems became too complex to be solved optimally, or were not solvable at all in the given time limit.
This fact is also the reason for the large difference between GM and RG which could not be observed for the (much simpler) BUC before: The higher complexity of the planning problem made the solutions that are found within the given time limit less optimal. 
Hence, the results show that the compilation process of the RG approach had a much higher impact onto the optimality of the solutions than the transformation procedure of the GM approach.
Through a detailed analysis of the generated plans, we found that this is mainly caused by the fact that, in contrast to the GM approach, the RG approach changes the structure of the actions space of a planning problem in a way that most planning heuristics are not able to deal with.

As the GM approach provided a much better overall performance, in all following experiments, we only considered the GM approach.

\paragraph*{Data-Driven Goal Recognition Results}
\label{subsec:comparisonDataDrivenResults}
\input{figures/figures_tex/evaluationFigure_comparison_dataDriven}
Next, we compare the different data-driven goal recognition methods.
Figure \ref{fig:CMUDataDrivenComparison} shows the average, cross-validated goal recognition accuracies of the NBM, KNN, and XGBoost for the CMU dataset. 
As the LSTM approach did not achieve accuracy values above 25\% for any training set size, we did not include the results in Figure \ref{fig:CMUDataDrivenComparison}.
For KNN and XGBoost, we compare performances of the fluent-based and action-based data encodings, as introduced in Section \ref{subsec:goal-rec-exp-design}.

The results show that all approaches performed much better, especially early in the observation sequence, when the planning state-based data encoding was used.
This shows that, in case of the CMU domain, the symbolic planning states encode more useful information regarding the actual goal of an observed agent than the sequence of observed actions.
Furthermore, the accuracies of all three methods did not depend strongly on the amount of available training data.
Interesting to note is that even though the NBM is the model with the lowest computational complexity, it was still not outperformed by the (slightly) more complex KNN and XGBoost models.
Hence, overall, the NBM is the most favorable data-driven model for this scenario, especially in mobile computing scenarios, where computational efficiency is of high relevance.

\subsection{Hybrid Goal Recognition}
\label{subsec:extendedCMUResults}
\input{figures/figures_tex/evaluationFigure_CMU_extended}
In this section, we assess the performance of the hybrid goal recognition models (i.e., Weighted Sum (WS) and Tiebreaking (TB)) in comparison to the purely data-driven NBM approach and the purely planning-based GM method. 
Figure \ref{fig:CMUHybridEvaluation} shows the average, cross-validated goal recognition accuracies of these approaches for the CMU dataset.

The results show that both hybrid approaches were at least as good as the GM and NBM approaches for small training set sizes.
For larger training set sizes (i.e., $n \geq 3$), the TB approach was increasingly outperformed by the NBM early in the recognition process (i.e., when only a small fraction of the observations were seen).
The reason for this is that the TB approach relies strongly on the predictions of the GM approach, which also became increasingly outperformed by the NBM early in the observation sequence with increasing training set sizes.
In contrast, the WS approach was not outperformed by the NBM, but reached at least similar performance as the NBM also when only a small fraction of the observations were seen.
The WS approach was even able to substantially outperform both the NBM and the GM approaches early in the observation sequences when, depending on the training set size, between 3\% and 30\% of the observations were used.
This effect was most prominent when training set sizes between $n=3$ and $n=7$ were used.

The results show that for $n \geq 3$, the planning-based and data-driven methods complemented each other well regarding recognition performance.
While the NBM approach achieved the best performances early in the observation sequences (i.e., less than 10\% - 20\% of the observations), the GM approach outperformed the NBM later in the  observation sequences (i.e., more than 10\% - 20\% of the observations). 
The hybrid WS approach was able to leverage on the strengths of the two individual approaches, constantly performing as good or better as each of them.

\subsection{Scalability of Hybrid Goal Recognition}
\label{subsec:artificiallyExtendedResults}

\paragraph*{Evaluating Scalability on Extended Real-World Dataset}
\input{figures/figures_tex/evaluationFigure_CMU_artificial}
Next, we investigate the scalability of the methods, by assessing goal recognition performance when the number of goals is increased. 
Figure \ref{fig:CMUArtificialHybridEvaluation} shows the cross-valiated mean accuracy of the GM, NBM, TB, and WS approaches for the ACMU dataset (i.e., with sampled observation sequences). 

Due to the doubled number of goals, the GM and the NBM approaches achieve a significantly lower recognition accuracy compared to the results for the CMU dataset that has not been extended with artificial data.
Nevertheless, it can be observed that the recognition performance of the GM approach converges towards the GM performance on the not artificially extended CMU dataset with an increasing fraction of observations that were used for recognition.
The results also show that the NBM approach, even when only a small number of training examples were used (i.e., $n=3$), is able to achieve a better goal recognition performance than the GM when less than 5\% of the observation sequences were seen.
In addition, the results show that the WS approach again performs similarly well or better than the two individual approaches.

The differences between the achieved recognition performances of the GM and NBM approaches are even larger early and late in the observation sequences than for the standard CMU dataset.
This indicates that increasing the number of possible goals makes the weaknesses of the individual approaches even more prominent and hence, using a hybrid approach that is able to compensate for them is even more favorable.
Note that this observation only holds when a limited number of training data is available as the performance of the NBM naturally will increase when more training data is available.
Nevertheless, it is very common in practice that annotated training examples are scarce as manually annotating observation sequences is costly and error-prone.

In summary, the results show that the hybrid recognition approach still achieves good goal recognition performance when the number of possible goals increases.
Moreover, the results indicate that using a hybrid approach is even more beneficial when the number of goals increases, compared to purely data-driven or purely planning-based methods.

\paragraph*{Evaluating Scalability on an Artificial Dataset}
\input{figures/figures_tex/evaluationFigureLogistics}
Finally, we further investigate the scalability of the methods by applying them to a benchmark plan recognition domain (which has simpler plans, but more possible goals than the real-world CMU domain). 
Figure \ref{fig:ComparisonArtificialDomains} shows the mean goal recognition accuracy of the GM, NBM, TB, and WS approaches for the logistics planning domain.
As for the CMU domain, the NBM performed better than the GM approach early in the observation sequences (i.e., when less than, depending on $n$, 5\% - 20\% of the observations were seen) and the GM performed better later in the observation sequences.
However, for the logistics domain, the NBM only achieved slightly better performance than the GM approach.

Interestingly, in contrast to the experiments with the CMU Dataset, the Tiebreaking (TB) approach also constantly performed as good or better than the two individual approaches (in addition to WS, as for the CMU dataset).
The main reason for this behavior is the fact that the assumptions underlying the TB approach hold more firmly for the LOG domain: 
TB assumes that the planning-based approach (GM, in this case) never predicts a wrong goal to be most probable, but only predicts multiple, equally likely goals (one of which is correct). 
This assumption only holds if the involved plans are  optimal.
The logistics domain, however, has substantially lower complexity than the CMU domain, such that the MetricFF planner was able to find more optimal plans in the given time limit. 
Thus, the assumptions of TB hold and TB could achieve better results than for the CMU and ACMU datasets.
In summary, the results show that our hybrid goal recognition approach is also beneficial in artificial planning domains where the number of goals is substantially higher than in the investigated real-world domain.


\section{Related Work} \label{sec:RelatedWork}
Existing approaches to goal- and plan recognition can be divided into model-based and model-free approaches.
Model-based approaches typically reason over handcrafted symbolic domain models to solve the recognition task.
In contrast, model-free approaches treat the recognition problem as a classification problem and learn to predict the current user goal from data and, thus, are data-driven.

Early model-based approaches to plan recognition relied on complete plan libraries that encode possible user behavior to recognize the current plan from observed user actions \cite{kautz1986generalized,charniak1993bayesian}.
However, these approaches require a large manual modeling effort, which is infeasible in large domains.
To overcome this issue, a new class of approaches to plan recognition that no longer required complete plan libraries, but only a domain model that defines possible states and actions, was proposed.
The PRAP approaches considered in this work \cite{ramirez2010probabilistic,pereira2020landmark}, \cite{vered2016mirroring} belong to this class.
Another example approach that relies on the use of classical planning systems is the approach by Sohrabi \textit{et al.} \cite{sohrabi2016revisited}.
They propose to use a top-k planner to generate the top-k plans for all possible goals in order to obtain which goal a user currently intents to achieve.
Nevertheless, most of these approaches have, so far, only been evaluated on relatively small, artificial domains, and hence, it is not clear whether they are also applicable to real-world scenarios.
We have shown that these PRAP approaches indeed show good performance in a real-world setting, but have problems in capturing relations between observations and user goals that cannot be properly modeled manually.
Some other recent approaches to goal recognition in smart environments also belong to this class of approaches \cite{yordanova2017s,yordanova2019analysing}.
Consequently, they have the same problems as the approaches considered in this work.

In contrast, model-free approaches learn to predict the most probable user goal directly from data.
Hence, they have the potential to learn the relations between actions and user goals that are not properly captured by model-based approaches.
In \cite{jameson_towards_1997}, the authors propose to use a BN model to predict the current quest of an observed player of a computer game.
Recently,  approaches that applied deep learning methods to goal recognition problems have been proposed \cite{min2016player,amado2018lstm}.
For example, Min \textit{et al.} \cite{min2016player} applied a LSTM for player goal recognition in digital games.
However, model-free approaches usually require large amounts of training data to produce reasonable results.
In the case of deep learning models, several thousands of annotated training examples are required to train the model adequately.
Such amounts of training data are usually not easily available for real-world scenarios.
Regarding this aspect, model-based approaches have a clear advantage because they can rely on handcrafted domain knowledge.
Thus, to benefit from both paradigms' strengths, we propose a hybrid approach that combines a model-based and a model-free method.


\section{Conclusion and Future Work} \label{sec:conclusion}
In this work, we investigated whether existing plan recognition as planning (PRAP) approaches can be applied to solve the online goal recognition problem in a real-world kitchen scenario.
More explicitly, we conducted several empirical goal recognition experiments on the basis of the well-known CMU Kitchen Dataset, which contains observation sequences for five possible goals of up to 36 different subjects.
We found that such PRAP approaches can indeed be used to solve the online goal recognition problems in real-world scenarios.
Nevertheless, we also revealed and analyzed some major limitations of PRAP approaches when applied to such scenarios.
As a possible solution, we proposed a hybrid goal recognition method, which combines a symbolic PRAP approach and a data-driven model.
We showed that the hybrid approach is able to recognize an agent's true goal more reliably than the PRAP approaches, especially early in an observation sequence (i.e., when only a small fraction of the observations were seen).
To investigate the scalability of the proposed hybrid approach in terms of the number of possible goals, we conducted an experiment based on an artificially extended version of the CMU Kitchen Dataset.
The results of these experiments indicate that the advantages of using a hybrid approach are becoming even more prominent with an increasing number of possible goals.

In summary, we showed that using a hybrid goal recognition method provides a valuable improvement compared to state-of-the-art purely symbolic and data-driven goal recognition methods.
It was found that our proposed hybrid method is able to outperform purely symbolic and data-driven methods and recognize the correct goal more reliably based on a lower number of observations, although only a small number of training examples are used.
This result substantially improves the usefulness of goal recognition for intelligent assistance systems, as recognizing a goal early opens much more possibilities for supportive reactions of the system.
Furthermore, it is usually very expansive to obtain annotated training examples for real-world application scenarios.
Hence, being able to provide valuable results based on limited numbers of training examples is an important requirement for potential goal recognition methods that should be applied to real-world application scenarios.
Nevertheless, we still see some potential for improvements of the extended approach in future work.
One direction is to optimize the procedure that is used to combine the results of the two individual approaches.
Another direction that we plan to investigate in future work is to use more complex tractable probabilistic models, like Sum-Product Networks \cite{poon2011sum}.

\section{Acknowledgements}
The data used in this paper was obtained from kitchen.cs.cmu.edu and the data collection was funded in part by the National Science Foundation [grant number EEEC-0540865].
This work was supported by the German Federal Ministry of Education and Research (BMBF) [grant number 01lS18079C].

\appendix
\section{Sampling Artificial Observation Sequences}
\label{sec:appendixSampling}
Algorithm \ref{alg:sequenceSampling} summarizes the sampling procedure that is used in this work to sample artificial observation sequences.
Here, $p_{planAction}$ is a parameter that specifies the \emph{goal-directedness}, i.e., the probability that the next action is taken from a precomputed optimal plan.
When this is not the case, the next action is sampled out of the set of actions that are currently applicable in the current planning state.
These actions are randomly drawn from all actions that are applicable in a certain state of the planning domain, following two predefined probability models that model the probability that an interaction with a certain object $O$ is observed given that we want to reach a goal $G$ ($P(O|G)$), and the probability that a certain kind of action $A$ is observed given that we want to reach goal $G$ ($P(A|G)$).

The distribution $P(A|G, S)$ that is used to sample an action at random is defined as follows:
\begin{equation}
P(A=a_i | g, s) \propto 
\begin{cases}
w_i & \text{ if $a_i$ applicable in s}\\
0 & \text{ otherwise}
\end{cases}
\end{equation}
That is, only applicable actions can be selected.
The weight $w_i$ of an action depends on the corresponding ``action type'' $AT(a_i)$ and the set of objects $OB(a_i)$ with that each action interacts.
The underlying intuition is the observation that depending on the current goal, the agent will choose actions of different action types with higher probabilities than others and also interact with certain objects with higher probabilities.
Based on this intuition, we use randomly initialized weight score distributions $W(AT|G)$ and $W(OB|G)$ to determine the weights $w_i$ via
\begin{equation}
w_i = W(AT(a_i)|g) \prod_{x \in OB(a_i)}{W(x|g)},
\end{equation}

\begin{algorithm}[tb]
\caption{Sample artificial observation sequence for goal $g$.}
\label{alg:sequenceSampling}
\begin{algorithmic}
\STATE sampledPlan $\leftarrow$ ()
\STATE cState $\leftarrow$ initialPlanningState
\STATE optPlan $\leftarrow$ computeOptimalPlan(cState, g)
\STATE $i = 0$

\WHILE{\textit{goal not reached}}
	\STATE $r$ $\leftarrow$ random(0, 1)
	\IF{$r < p_{planAction}$}
		\STATE sAction $\leftarrow$ optPlan.getAction(i)
		\STATE cState $\leftarrow$ cState.apply(sAction)
		\STATE $i \leftarrow i+1$
	\ELSE
		\STATE sAction $\leftarrow$ sampleActionFromApplicableActions(cState)
		\STATE cState $\leftarrow$ cState.apply(sAction)
		\STATE optPlan $\leftarrow$ computeOptimalPlan(cState, g)
		\STATE $i = 0$
	\ENDIF
	\STATE sampledPlan $\leftarrow$ concat(sampledPlan, sAction)
\ENDWHILE

\end{algorithmic}
\end{algorithm}

We initialize the parameters of $P(A|G)$ and $P(O|G)$ randomly and use the MetricFF planner to compute the initial plan.
Each time an action from the sampling model is selected, the optimal plan from the resulting state is recomputed.

\vskip 0.2in
\bibliography{hybridGoalRecognition}
\bibliographystyle{theapa}

\end{document}

%% file: figures/figures_tex/evaluationFigure_RG_GM_comparison.tex
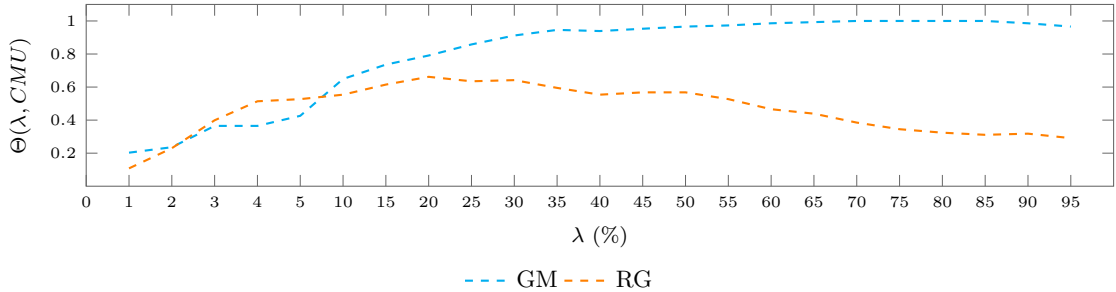
\begin{figure}[t]
    \centering
        \subfloat{
            \begin{tikzpicture}
            \pgfplotsset{every x tick label/.append style={font=\tiny}}
            \pgfplotsset{every y tick label/.append style={font=\tiny}}
                \begin{axis}[
                    width=1\linewidth,
                    height=4cm,
                    xlabel={\scriptsize{$\lambda$ (\%)}},
                    ylabel={\scriptsize{$\Theta(\lambda,CMU)$}},
                    xlabel near ticks,
                    ylabel near ticks,
                    xmin=0, xmax=24,
                    ymin=0, ymax=1.1,
                    xtick={0,1,2,3,4,5,6,7,8,9,10,11,12,13,14,15,16,17,18,19,20,21,22,23},
                    xticklabels={0,1,2,3,4,5,10,15,20,25,30,35,40,45,50,55,60,65,70,75,80,85,90,95},
                    ytick={0.2,0.4,0.6,0.8,1},
                    legend pos=north west,
                    ymajorgrids=false,
                    xmajorgrids=false,
                    major grid style={line width=.1pt,draw=gray!50},
                    x axis line style={draw=black!60},
                    tick style={draw=black!60},
                    legend columns=3,
                    legend style={draw=none},
                    legend entries={\footnotesize{GM},\footnotesize{RG}},
                    legend to name={plotLabel4}
                ]

                \addplot[
                    color=cyan,
	                mark=none,
	                dashed,
	                thick
	               ]
	                coordinates { (1.0,0.203)(2.0,0.236)(3.0,0.365)(4.0,0.365)(5.0,0.426)(6.0,0.649)(7.0,0.736)(8.0,0.791)(9.0,0.858)(10.0,0.912)(11.0,0.946)(12.0,0.939)(13.0,0.953)(14.0,0.966)(15.0,0.973)(16.0,0.986)(17.0,0.993)(18.0,1.0)(19.0,1.0)(20.0,1.0)(21.0,1.0)(22.0,0.986)(23.0,0.966)};
	           \addplot[
                    color=orange,
	                mark=none,
	                dashed,
	                thick,
	               ]
	                coordinates { (1.0,0.108)(2.0,0.23)(3.0,0.399)(4.0,0.514)(5.0,0.527)(6.0,0.554)(7.0,0.615)(8.0,0.662)(9.0,0.635)(10.0,0.642)(11.0,0.595)(12.0,0.554)(13.0,0.568)(14.0,0.568)(15.0,0.527)(16.0,0.466)(17.0,0.439)(18.0,0.385)(19.0,0.345)(20.0,0.324)(21.0,0.311)(22.0,0.318)(23.0,0.291)};

                \end{axis}
            \end{tikzpicture}}
            \newline
            \vspace{-0.2cm}
    \ref{plotLabel4}\vspace{-0.3cm}
    \caption{Mean accuracy of the planning-based methods (RG and GM) on the \emph{CMU Dataset without artifical samples}.
    }
    \label{fig:CMURGGMComparison}
\end{figure}

%% file: figures/figures_tex/evaluationFigure_comparison_dataDriven.tex
\begin{figure*}[t]
    \centering
        \subfloat{
            \begin{tikzpicture}
            \pgfplotsset{every x tick label/.append style={font=\tiny}}
            \pgfplotsset{every y tick label/.append style={font=\tiny}}
                \begin{axis}[
                    width=0.55\linewidth,
                    height=4cm,
                    title={\footnotesize{n=1}},
                    title style={at={(0.063,0.915)},anchor=north,draw=black!60,fill=white},
                    ylabel={\scriptsize{$Acc(\lambda,CMU)$}},
                    xlabel near ticks,
                    ylabel near ticks,
                    xmin=0, xmax=24,
                    ymin=0, ymax=1.1,
                    xtick={0,1,2,3,4,5,6,7,8,9,10,11,12,13,14,15,16,17,18,19,20,21,22,23},
                    xticklabels={0,1,2,3,4,5,10,15,20,25,30,35,40,45,50,55,60,65,70,75,80,85,90,95},
                    ytick={0.2,0.4,0.6,0.8,1},
                    legend pos=north west,
                    ymajorgrids=false,
                    xmajorgrids=false,
                    major grid style={line width=.1pt,draw=gray!50},
                    x axis line style={draw=black!60},
                    tick style={draw=black!60},
                    legend columns=5,
                    legend style={draw=none},
                    legend entries={\footnotesize{NBM},\footnotesize{XGBoost (actions)},\footnotesize{XGBoost (states)},\footnotesize{KNN (actions)},\footnotesize{KNN (states)}},
                    legend to name={plotLabel3}
                ]

                \addplot[
					color=green!70,
					mark=none,
					dashdotted,
					thick
					]
					coordinates { (1.0,0.197)(2.0,0.197)(3.0,0.197)(4.0,0.197)(5.0,0.197)(6.0,0.197)(7.0,0.197)(8.0,0.197)(9.0,0.197)(10.0,0.197)(11.0,0.197)(12.0,0.197)(13.0,0.197)(14.0,0.197)(15.0,0.197)(16.0,0.197)(17.0,0.197)(18.0,0.197)(19.0,0.197)(20.0,0.197)(21.0,0.197)(22.0,0.197)(23.0,0.197)};
                \addplot[
					color=lightgray,
					mark=none,
					dashed,
					thick
					]
					coordinates { (1.0,0.197)(2.0,0.197)(3.0,0.197)(4.0,0.197)(5.0,0.197)(6.0,0.197)(7.0,0.197)(8.0,0.197)(9.0,0.197)(10.0,0.197)(11.0,0.197)(12.0,0.197)(13.0,0.197)(14.0,0.197)(15.0,0.197)(16.0,0.197)(17.0,0.197)(18.0,0.197)(19.0,0.197)(20.0,0.197)(21.0,0.197)};
				\addplot[
					color=lightgray,
					mark=none,
					densely dotted,
					thick
					]
					coordinates { (1.0,0.197)(2.0,0.197)(3.0,0.197)(4.0,0.197)(5.0,0.197)(6.0,0.197)(7.0,0.197)(8.0,0.197)(9.0,0.197)(10.0,0.197)(11.0,0.197)(12.0,0.197)(13.0,0.197)(14.0,0.197)(15.0,0.197)(16.0,0.197)(17.0,0.197)(18.0,0.197)(19.0,0.197)(20.0,0.197)(21.0,0.197)};
				\addplot[
					color=black!90,
					mark=none,
					dashed,
					thick
					]
					coordinates { (1.0,0.197)(2.0,0.197)(3.0,0.197)(4.0,0.197)(5.0,0.197)(6.0,0.197)(7.0,0.197)(8.0,0.197)(9.0,0.197)(10.0,0.197)(11.0,0.197)(12.0,0.197)(13.0,0.197)(14.0,0.197)(15.0,0.197)(16.0,0.197)(17.0,0.197)(18.0,0.197)(19.0,0.197)(20.0,0.197)(21.0,0.197)};
				\addplot[
					color=black!70,
					mark=none,
					densely dotted,
					thick
					]
					coordinates { (1.0,0.197)(2.0,0.197)(3.0,0.197)(4.0,0.197)(5.0,0.197)(6.0,0.197)(7.0,0.197)(8.0,0.197)(9.0,0.197)(10.0,0.197)(11.0,0.197)(12.0,0.197)(13.0,0.197)(14.0,0.197)(15.0,0.197)(16.0,0.197)(17.0,0.197)(18.0,0.197)(19.0,0.197)(20.0,0.197)(21.0,0.197)};

                \end{axis}
            \end{tikzpicture}}
        \subfloat{
             \begin{tikzpicture}
            \pgfplotsset{every x tick label/.append style={font=\tiny}}
            \pgfplotsset{every y tick label/.append style={font=\tiny}}
                \begin{axis}[
                    width=0.55\linewidth,
                    height=4cm,
                    title={\footnotesize{n=3}},
                     title style={at={(0.063,0.915)},anchor=north,draw=black!60,fill=white},
                    xlabel near ticks,
                    ylabel near ticks,
                    xmin=0, xmax=24,
                    ymin=0, ymax=1.1,
                    xtick={0,1,2,3,4,5,6,7,8,9,10,11,12,13,14,15,16,17,18,19,20,21,22,23},
                    xticklabels={0,1,2,3,4,5,10,15,20,25,30,35,40,45,50,55,60,65,70,75,80,85,90,95},
                    ytick={0.2,0.4,0.6,0.8,1},
                    legend pos=north west,
                    ymajorgrids=false,
                    xmajorgrids=false,
                    major grid style={line width=.1pt,draw=gray!50},
                    x axis line style={draw=black!60},
                    tick style={draw=black!60},
                ]

                \addplot[
					color=green!70,
					mark=none,
					dashdotted,
					thick
					]
					coordinates { (1.0,0.394)(2.0,0.4)(3.0,0.423)(4.0,0.438)(5.0,0.45)(6.0,0.471)(7.0,0.479)(8.0,0.486)(9.0,0.491)(10.0,0.493)(11.0,0.494)(12.0,0.494)(13.0,0.494)(14.0,0.493)(15.0,0.494)(16.0,0.494)(17.0,0.495)(18.0,0.495)(19.0,0.495)(20.0,0.495)(21.0,0.495)(22.0,0.496)(23.0,0.496)};
                \addplot[
					color=lightgray,
					mark=none,
					dashed,
					thick
					]
					coordinates { (1.0,0.198)(2.0,0.213)(3.0,0.236)(4.0,0.259)(5.0,0.278)(6.0,0.339)(7.0,0.385)(8.0,0.401)(9.0,0.416)(10.0,0.424)(11.0,0.433)(12.0,0.438)(13.0,0.443)(14.0,0.446)(15.0,0.449)(16.0,0.452)(17.0,0.454)(18.0,0.454)(19.0,0.456)(20.0,0.456)(21.0,0.456)(22.0,0.456)(23.0,0.459)};
				\addplot[
					color=lightgray,
					mark=none,
					densely dotted,
					thick
					]
					coordinates { (1.0,0.398)(2.0,0.404)(3.0,0.422)(4.0,0.432)(5.0,0.439)(6.0,0.454)(7.0,0.464)(8.0,0.47)(9.0,0.471)(10.0,0.475)(11.0,0.478)(12.0,0.479)(13.0,0.48)(14.0,0.482)(15.0,0.483)(16.0,0.483)(17.0,0.482)(18.0,0.484)(19.0,0.482)(20.0,0.482)(21.0,0.483)(22.0,0.482)(23.0,0.477)};
				\addplot[
					color=black!90,
					mark=none,
					dashed,
					thick
					]
					coordinates { (1.0,0.197)(2.0,0.205)(3.0,0.213)(4.0,0.222)(5.0,0.229)(6.0,0.29)(7.0,0.329)(8.0,0.351)(9.0,0.374)(10.0,0.393)(11.0,0.407)(12.0,0.416)(13.0,0.424)(14.0,0.429)(15.0,0.434)(16.0,0.437)(17.0,0.44)(18.0,0.441)(19.0,0.442)(20.0,0.442)(21.0,0.442)};
				\addplot[
					color=black!70,
					mark=none,
					densely dotted,
					thick
					]
					coordinates { (1.0,0.387)(2.0,0.382)(3.0,0.386)(4.0,0.392)(5.0,0.407)(6.0,0.428)(7.0,0.439)(8.0,0.448)(9.0,0.452)(10.0,0.457)(11.0,0.46)(12.0,0.463)(13.0,0.465)(14.0,0.467)(15.0,0.469)(16.0,0.47)(17.0,0.47)(18.0,0.472)(19.0,0.471)(20.0,0.471)(21.0,0.469)(22.0,0.471)(23.0,0.468)};

                \end{axis}
            \end{tikzpicture}}
        \vspace{-0.25cm}
        \subfloat{
             \begin{tikzpicture}
            \pgfplotsset{every x tick label/.append style={font=\tiny}}
            \pgfplotsset{every y tick label/.append style={font=\tiny}}
                \begin{axis}[
                    width=0.55\linewidth,
                    height=4cm,
                    title={\footnotesize{n=5}},
                     title style={at={(0.063,0.915)},anchor=north,draw=black!60,fill=white},
                    ylabel={\scriptsize{$Acc(\lambda,CMU)$}},
                    xlabel near ticks,
                    ylabel near ticks,
                    xmin=0, xmax=24,
                    ymin=0, ymax=1.1,
                    xtick={0,1,2,3,4,5,6,7,8,9,10,11,12,13,14,15,16,17,18,19,20,21,22,23},
                    xticklabels={0,1,2,3,4,5,10,15,20,25,30,35,40,45,50,55,60,65,70,75,80,85,90,95},
                    ytick={0.2,0.4,0.6,0.8,1},
                    legend pos=north west,
                    ymajorgrids=false,
                    xmajorgrids=false,
                    major grid style={line width=.1pt,draw=gray!50},
                    x axis line style={draw=black!60},
                    tick style={draw=black!60},
                ]

                \addplot[
					color=green!70,
					mark=none,
					dashdotted,
					thick
					]
					coordinates { (1.0,0.459)(2.0,0.467)(3.0,0.51)(4.0,0.53)(5.0,0.542)(6.0,0.569)(7.0,0.582)(8.0,0.591)(9.0,0.593)(10.0,0.6)(11.0,0.606)(12.0,0.604)(13.0,0.604)(14.0,0.606)(15.0,0.607)(16.0,0.609)(17.0,0.61)(18.0,0.612)(19.0,0.612)(20.0,0.611)(21.0,0.612)(22.0,0.612)(23.0,0.611)};
                \addplot[
					color=lightgray,
					mark=none,
					dashed,
					thick
					]
					coordinates { (1.0,0.201)(2.0,0.214)(3.0,0.247)(4.0,0.287)(5.0,0.322)(6.0,0.42)(7.0,0.486)(8.0,0.521)(9.0,0.547)(10.0,0.569)(11.0,0.587)(12.0,0.592)(13.0,0.599)(14.0,0.604)(15.0,0.611)(16.0,0.615)(17.0,0.617)(18.0,0.618)(19.0,0.62)(20.0,0.62)(21.0,0.62)(22.0,0.62)(23.0,0.622)};
				\addplot[
					color=lightgray,
					mark=none,
					densely dotted,
					thick
					]
					coordinates { (1.0,0.506)(2.0,0.512)(3.0,0.536)(4.0,0.552)(5.0,0.561)(6.0,0.598)(7.0,0.613)(8.0,0.624)(9.0,0.626)(10.0,0.632)(11.0,0.637)(12.0,0.641)(13.0,0.644)(14.0,0.647)(15.0,0.651)(16.0,0.65)(17.0,0.651)(18.0,0.65)(19.0,0.652)(20.0,0.65)(21.0,0.65)(22.0,0.646)(23.0,0.637)};
				\addplot[
					color=black!90,
					mark=none,
					dashed,
					thick
					]
					coordinates { (1.0,0.209)(2.0,0.226)(3.0,0.244)(4.0,0.259)(5.0,0.276)(6.0,0.387)(7.0,0.452)(8.0,0.501)(9.0,0.53)(10.0,0.56)(11.0,0.584)(12.0,0.599)(13.0,0.614)(14.0,0.621)(15.0,0.63)(16.0,0.635)(17.0,0.639)(18.0,0.641)(19.0,0.642)(20.0,0.642)(21.0,0.642)};
				\addplot[
					color=black!70,
					mark=none,
					densely dotted,
					thick
					]
					coordinates { (1.0,0.472)(2.0,0.462)(3.0,0.48)(4.0,0.497)(5.0,0.519)(6.0,0.56)(7.0,0.581)(8.0,0.60)(9.0,0.61)(10.0,0.618)(11.0,0.627)(12.0,0.631)(13.0,0.633)(14.0,0.638)(15.0,0.644)(16.0,0.645)(17.0,0.647)(18.0,0.651)(19.0,0.65)(20.0,0.647)(21.0,0.645)(22.0,0.648)(23.0,0.641)};

                \end{axis}
            \end{tikzpicture}}
        \subfloat{
             \begin{tikzpicture}
            \pgfplotsset{every x tick label/.append style={font=\tiny}}
            \pgfplotsset{every y tick label/.append style={font=\tiny}}
                \begin{axis}[
                    width=0.55\linewidth,
                    height=4cm,
                    title={\footnotesize{n=7}},
                    title style={at={(0.063,0.915)},anchor=north,draw=black!60,fill=white},
                    xlabel near ticks,
                    ylabel near ticks,
                    xmin=0, xmax=24,
                    ymin=0, ymax=1.1,
                    xtick={0,1,2,3,4,5,6,7,8,9,10,11,12,13,14,15,16,17,18,19,20,21,22,23},
                    xticklabels={0,1,2,3,4,5,10,15,20,25,30,35,40,45,50,55,60,65,70,75,80,85,90,95},
                    ytick={0.2,0.4,0.6,0.8,1},
                    legend pos=north west,
                    ymajorgrids=false,
                    xmajorgrids=false,
                    major grid style={line width=.1pt,draw=gray!50},
                    x axis line style={draw=black!60},
                    tick style={draw=black!60},
                ]

                \addplot[
					color=green!70,
					mark=none,
					dashdotted,
					thick
					]
					coordinates { (1.0,0.51)(2.0,0.524)(3.0,0.589)(4.0,0.622)(5.0,0.642)(6.0,0.703)(7.0,0.726)(8.0,0.743)(9.0,0.751)(10.0,0.759)(11.0,0.764)(12.0,0.764)(13.0,0.764)(14.0,0.765)(15.0,0.765)(16.0,0.769)(17.0,0.771)(18.0,0.772)(19.0,0.771)(20.0,0.772)(21.0,0.773)(22.0,0.773)(23.0,0.772)};
                \addplot[
					color=lightgray,
					mark=none,
					dashed,
					thick
					]
					coordinates { (1.0,0.198)(2.0,0.225)(3.0,0.269)(4.0,0.317)(5.0,0.361)(6.0,0.473)(7.0,0.555)(8.0,0.602)(9.0,0.635)(10.0,0.66)(11.0,0.682)(12.0,0.688)(13.0,0.695)(14.0,0.702)(15.0,0.706)(16.0,0.71)(17.0,0.712)(18.0,0.715)(19.0,0.719)(20.0,0.719)(21.0,0.72)(22.0,0.72)(23.0,0.721)};
				\addplot[
					color=lightgray,
					mark=none,
					densely dotted,
					thick
					]
					coordinates { (1.0,0.555)(2.0,0.571)(3.0,0.602)(4.0,0.627)(5.0,0.644)(6.0,0.691)(7.0,0.715)(8.0,0.727)(9.0,0.732)(10.0,0.741)(11.0,0.748)(12.0,0.753)(13.0,0.753)(14.0,0.759)(15.0,0.761)(16.0,0.762)(17.0,0.761)(18.0,0.762)(19.0,0.763)(20.0,0.761)(21.0,0.763)(22.0,0.756)(23.0,0.745)};
				\addplot[
					color=black!90,
					mark=none,
					dashed,
					thick
					]
					coordinates { (1.0,0.21)(2.0,0.227)(3.0,0.256)(4.0,0.283)(5.0,0.304)(6.0,0.425)(7.0,0.517)(8.0,0.578)(9.0,0.616)(10.0,0.65)(11.0,0.677)(12.0,0.691)(13.0,0.709)(14.0,0.724)(15.0,0.744)(16.0,0.752)(17.0,0.759)(18.0,0.764)(19.0,0.766)(20.0,0.766)(21.0,0.766)(22.0,0.766)(23.0,0.766)};
				\addplot[
					color=black!70,
					mark=none,
					densely dotted,
					thick
					]
					coordinates { (1.0,0.538)(2.0,0.532)(3.0,0.548)(4.0,0.572)(5.0,0.606)(6.0,0.644)(7.0,0.681)(8.0,0.707)(9.0,0.718)(10.0,0.726)(11.0,0.736)(12.0,0.74)(13.0,0.742)(14.0,0.75)(15.0,0.759)(16.0,0.758)(17.0,0.763)(18.0,0.767)(19.0,0.766)(20.0,0.764)(21.0,0.763)(22.0,0.765)(23.0,0.761)};

                \end{axis}
            \end{tikzpicture}}
        \vspace{-0.25cm}
        \subfloat{
             \begin{tikzpicture}
            \pgfplotsset{every x tick label/.append style={font=\tiny}}
            \pgfplotsset{every y tick label/.append style={font=\tiny}}
                \begin{axis}[
                    width=0.55\linewidth,
                    height=4cm,
                    title={\footnotesize{n=9}},
                     title style={at={(0.063,0.915)},anchor=north,draw=black!60,fill=white},
                    xlabel={\scriptsize{$\lambda$ (\%)}},
                    ylabel={\scriptsize{$Acc(\lambda,CMU)$}},
                    xlabel near ticks,
                    ylabel near ticks,
                    xmin=0, xmax=24,
                    ymin=0, ymax=1.1,
                    xtick={0,1,2,3,4,5,6,7,8,9,10,11,12,13,14,15,16,17,18,19,20,21,22,23},
                    xticklabels={0,1,2,3,4,5,10,15,20,25,30,35,40,45,50,55,60,65,70,75,80,85,90,95},
                    ytick={0.2,0.4,0.6,0.8,1},
                    legend pos=north west,
                    ymajorgrids=false,
                    xmajorgrids=false,
                    major grid style={line width=.1pt,draw=gray!50},
                    x axis line style={draw=black!60},
                    tick style={draw=black!60},
                ]

                \addplot[
					color=green!70,
					mark=none,
					dashdotted,
					thick
					]
					coordinates { (1.0,0.601)(2.0,0.614)(3.0,0.675)(4.0,0.718)(5.0,0.752)(6.0,0.807)(7.0,0.834)(8.0,0.85)(9.0,0.857)(10.0,0.865)(11.0,0.871)(12.0,0.871)(13.0,0.871)(14.0,0.874)(15.0,0.874)(16.0,0.876)(17.0,0.882)(18.0,0.883)(19.0,0.882)(20.0,0.882)(21.0,0.883)(22.0,0.884)(23.0,0.884)};
                \addplot[
					color=lightgray,
					mark=none,
					dashed,
					thick
					]
					coordinates { (1.0,0.20)(2.0,0.243)(3.0,0.297)(4.0,0.349)(5.0,0.399)(6.0,0.541)(7.0,0.628)(8.0,0.678)(9.0,0.708)(10.0,0.74)(11.0,0.759)(12.0,0.772)(13.0,0.783)(14.0,0.79)(15.0,0.797)(16.0,0.801)(17.0,0.804)(18.0,0.808)(19.0,0.811)(20.0,0.813)(21.0,0.813)(22.0,0.813)(23.0,0.815)};
				\addplot[
					color=lightgray,
					mark=none,
					densely dotted,
					thick
					]
					coordinates { (1.0,0.609)(2.0,0.623)(3.0,0.666)(4.0,0.69)(5.0,0.71)(6.0,0.757)(7.0,0.791)(8.0,0.803)(9.0,0.81)(10.0,0.824)(11.0,0.827)(12.0,0.831)(13.0,0.835)(14.0,0.841)(15.0,0.845)(16.0,0.846)(17.0,0.846)(18.0,0.85)(19.0,0.851)(20.0,0.852)(21.0,0.851)(22.0,0.847)(23.0,0.837)};
				\addplot[
					color=black!90,
					mark=none,
					dashed,
					thick
					]
					coordinates { (1.0,0.197)(2.0,0.217)(3.0,0.261)(4.0,0.296)(5.0,0.326)(6.0,0.473)(7.0,0.577)(8.0,0.635)(9.0,0.67)(10.0,0.705)(11.0,0.731)(12.0,0.731)(13.0,0.74)(14.0,0.76)(15.0,0.772)(16.0,0.789)(17.0,0.793)(18.0,0.795)(19.0,0.797)(20.0,0.797)(21.0,0.797)(22.0,0.797)(23.0,0.798)};
				\addplot[
					color=black!70,
					mark=none,
					densely dotted,
					thick
					]
					coordinates { (1.0,0.651)(2.0,0.634)(3.0,0.658)(4.0,0.682)(5.0,0.718)(6.0,0.754)(7.0,0.796)(8.0,0.824)(9.0,0.835)(10.0,0.847)(11.0,0.854)(12.0,0.863)(13.0,0.867)(14.0,0.877)(15.0,0.884)(16.0,0.889)(17.0,0.892)(18.0,0.895)(19.0,0.895)(20.0,0.895)(21.0,0.893)(22.0,0.895)(23.0,0.89)};

                \end{axis}
            \end{tikzpicture}}
        \subfloat{
             \begin{tikzpicture}
            \pgfplotsset{every x tick label/.append style={font=\tiny}}
            \pgfplotsset{every y tick label/.append style={font=\tiny}}
                \begin{axis}[
                    width=0.55\linewidth,
                    height=4cm,
                    title={\footnotesize{n=11}},
                    title style={at={(0.0745,0.915)},anchor=north,draw=black!60,fill=white},
                    xlabel={\scriptsize{$\lambda$ (\%)}},
                    xlabel near ticks,
                    ylabel near ticks,
                    xmin=0, xmax=24,
                    ymin=0, ymax=1.1,
                    xtick={0,1,2,3,4,5,6,7,8,9,10,11,12,13,14,15,16,17,18,19,20,21,22,23},
                    xticklabels={0,1,2,3,4,5,10,15,20,25,30,35,40,45,50,55,60,65,70,75,80,85,90,95},
                    ytick={0.2,0.4,0.6,0.8,1},
                    legend pos=north west,
                    ymajorgrids=false,
                    xmajorgrids=false,
                    major grid style={line width=.1pt,draw=gray!50},
                    x axis line style={draw=black!60},
                    tick style={draw=black!60},
                ]

                \addplot[
					color=green!70,
					mark=none,
					dashdotted,
					thick
					]
					coordinates { (1.0,0.648)(2.0,0.657)(3.0,0.726)(4.0,0.755)(5.0,0.778)(6.0,0.829)(7.0,0.848)(8.0,0.855)(9.0,0.868)(10.0,0.876)(11.0,0.886)(12.0,0.884)(13.0,0.886)(14.0,0.885)(15.0,0.885)(16.0,0.886)(17.0,0.89)(18.0,0.889)(19.0,0.89)(20.0,0.889)(21.0,0.89)(22.0,0.893)(23.0,0.892)};
                \addplot[
					color=lightgray,
					mark=none,
					dashed,
					thick
					]
					coordinates { (1.0,0.198)(2.0,0.248)(3.0,0.316)(4.0,0.387)(5.0,0.438)(6.0,0.554)(7.0,0.641)(8.0,0.695)(9.0,0.732)(10.0,0.767)(11.0,0.793)(12.0,0.802)(13.0,0.812)(14.0,0.82)(15.0,0.829)(16.0,0.834)(17.0,0.836)(18.0,0.839)(19.0,0.846)(20.0,0.848)(21.0,0.848)(22.0,0.848)(23.0,0.851)};
				\addplot[
					color=lightgray,
					mark=none,
					densely dotted,
					thick
					]
					coordinates { (1.0,0.659)(2.0,0.669)(3.0,0.705)(4.0,0.735)(5.0,0.748)(6.0,0.796)(7.0,0.819)(8.0,0.832)(9.0,0.836)(10.0,0.851)(11.0,0.853)(12.0,0.855)(13.0,0.857)(14.0,0.86)(15.0,0.862)(16.0,0.864)(17.0,0.865)(18.0,0.868)(19.0,0.869)(20.0,0.869)(21.0,0.868)(22.0,0.867)(23.0,0.864)};
				\addplot[
					color=black!90,
					mark=none,
					dashed,
					thick
					]
					coordinates { (1.0,0.20)(2.0,0.231)(3.0,0.301)(4.0,0.356)(5.0,0.385)(6.0,0.519)(7.0,0.636)(8.0,0.687)(9.0,0.72)(10.0,0.749)(11.0,0.773)(12.0,0.788)(13.0,0.804)(14.0,0.814)(15.0,0.832)(16.0,0.84)(17.0,0.846)(18.0,0.849)(19.0,0.851)(20.0,0.851)(21.0,0.851)(22.0,0.851)(23.0,0.852)};
				\addplot[
					color=black!70,
					mark=none,
					densely dotted,
					thick
					]
					coordinates { (1.0,0.629)(2.0,0.622)(3.0,0.655)(4.0,0.685)(5.0,0.716)(6.0,0.75)(7.0,0.79)(8.0,0.814)(9.0,0.828)(10.0,0.835)(11.0,0.844)(12.0,0.85)(13.0,0.85)(14.0,0.858)(15.0,0.868)(16.0,0.875)(17.0,0.879)(18.0,0.883)(19.0,0.883)(20.0,0.883)(21.0,0.882)(22.0,0.885)(23.0,0.881)};
                \end{axis}
            \end{tikzpicture}}
            \vspace{-0.1cm}
    \ref{plotLabel3}\vspace{-0.3cm}
    \caption{Mean accuracy of the data-driven methods (NBM, KNN and XGBoost) on the \emph{CMU Dataset without additional samples} for different sizes of the training set $n$.}
    \label{fig:CMUDataDrivenComparison}
\end{figure*}
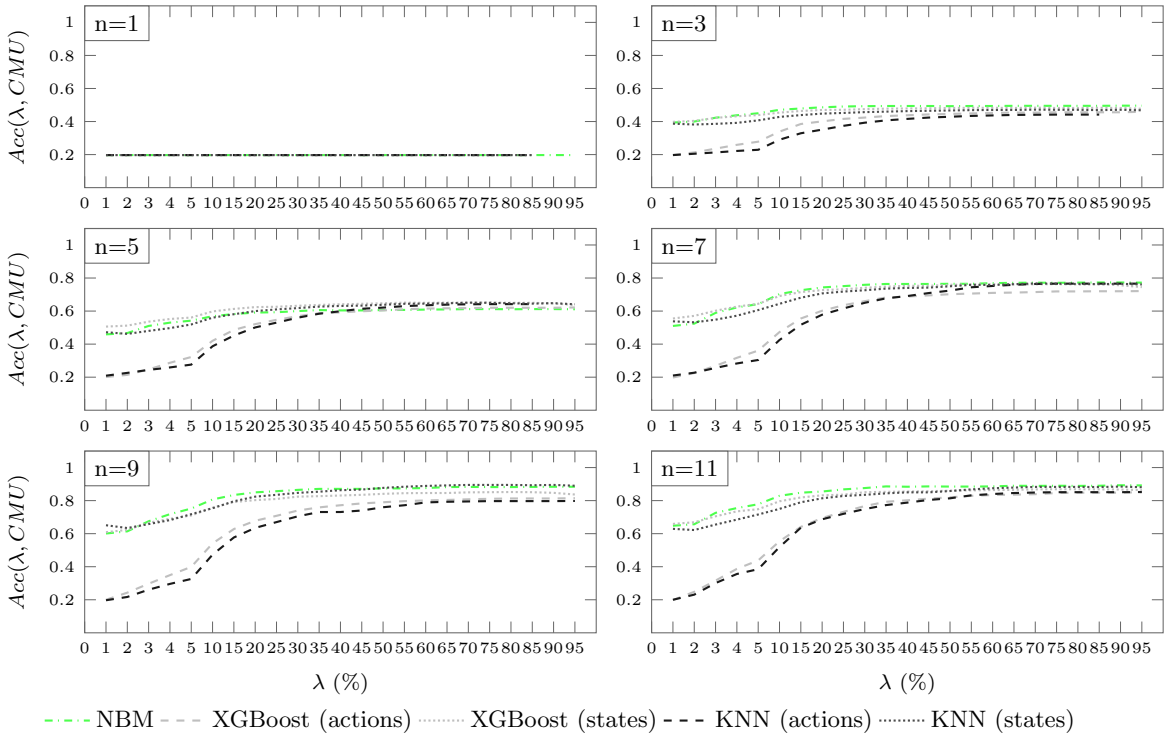

%% file: figures/figures_tex/evaluationFigure_CMU_extended.tex
\begin{figure*}[t]
    \centering
        \subfloat{
            \begin{tikzpicture}
            \pgfplotsset{every x tick label/.append style={font=\tiny}}
            \pgfplotsset{every y tick label/.append style={font=\tiny}}
                \begin{axis}[
                    width=0.55\linewidth,
                    height=4cm,
                    title={\footnotesize{n=1}},
                    title style={at={(0.063,0.915)},anchor=north,draw=black!60,fill=white},
                    ylabel={\scriptsize{$Acc(\lambda,CMU)$}},
                    xlabel near ticks,
                    ylabel near ticks,
                    xmin=0, xmax=24,
                    ymin=0, ymax=1.1,
                    xtick={0,1,2,3,4,5,6,7,8,9,10,11,12,13,14,15,16,17,18,19,20,21,22,23},
                    xticklabels={0,1,2,3,4,5,10,15,20,25,30,35,40,45,50,55,60,65,70,75,80,85,90,95},
                    ytick={0.2,0.4,0.6,0.8,1},
                    legend pos=north west,
                    ymajorgrids=false,
                    xmajorgrids=false,
                    major grid style={line width=.1pt,draw=gray!50},
                    x axis line style={draw=black!60},
                    tick style={draw=black!60},
                    legend columns=4,
                    legend style={draw=none},
                    legend entries={\footnotesize{GM},\footnotesize{NBM},\footnotesize{WS},\footnotesize{TB}},
                    legend to name={plotLabel2}
                ]

                \addplot[
                    color=cyan,
	                mark=none,
	                dashed,
	                thick
	               ]
	                coordinates { (1.0,0.203)(2.0,0.236)(3.0,0.365)(4.0,0.365)(5.0,0.426)(6.0,0.649)(7.0,0.736)(8.0,0.791)(9.0,0.858)(10.0,0.912)(11.0,0.946)(12.0,0.939)(13.0,0.953)(14.0,0.966)(15.0,0.973)(16.0,0.986)(17.0,0.993)(18.0,1.0)(19.0,1.0)(20.0,1.0)(21.0,1.0)(22.0,0.986)(23.0,0.966)};
                \addplot[
					color=green!70,
					mark=none,
					dashdotted,
					thick
					]
					coordinates { (1.0,0.197)(2.0,0.197)(3.0,0.197)(4.0,0.197)(5.0,0.197)(6.0,0.197)(7.0,0.197)(8.0,0.197)(9.0,0.197)(10.0,0.197)(11.0,0.197)(12.0,0.197)(13.0,0.197)(14.0,0.197)(15.0,0.197)(16.0,0.197)(17.0,0.197)(18.0,0.197)(19.0,0.197)(20.0,0.197)(21.0,0.197)(22.0,0.197)(23.0,0.197)};
                \addplot[
					color=red,
					mark=none,
					densely dotted,
					thick
					]
					coordinates { (1.0,0.187)(2.0,0.262)(3.0,0.405)(4.0,0.407)(5.0,0.48)(6.0,0.66)(7.0,0.753)(8.0,0.797)(9.0,0.867)(10.0,0.913)(11.0,0.946)(12.0,0.943)(13.0,0.953)(14.0,0.966)(15.0,0.973)(16.0,0.986)(17.0,0.993)(18.0,1.0)(19.0,1.0)(20.0,1.0)(21.0,1.0)(22.0,0.986)(23.0,0.966)};
				\addplot[
					color=black,
					mark=none,
					densely dotted,
					thick
					]
					coordinates { (1.0,0.252)(2.0,0.271)(3.0,0.395)(4.0,0.392)(5.0,0.46)(6.0,0.655)(7.0,0.753)(8.0,0.796)(9.0,0.867)(10.0,0.913)(11.0,0.946)(12.0,0.943)(13.0,0.953)(14.0,0.966)(15.0,0.973)(16.0,0.986)(17.0,0.993)(18.0,1.0)(19.0,1.0)(20.0,1.0)(21.0,1.0)(22.0,0.986)(23.0,0.966)};

                \end{axis}
            \end{tikzpicture}}
        \subfloat{
             \begin{tikzpicture}
            \pgfplotsset{every x tick label/.append style={font=\tiny}}
            \pgfplotsset{every y tick label/.append style={font=\tiny}}
                \begin{axis}[
                    width=0.55\linewidth,
                    height=4cm,
                    title={\footnotesize{n=3}},
                     title style={at={(0.063,0.915)},anchor=north,draw=black!60,fill=white},
                    xlabel near ticks,
                    ylabel near ticks,
                    xmin=0, xmax=24,
                    ymin=0, ymax=1.1,
                    xtick={0,1,2,3,4,5,6,7,8,9,10,11,12,13,14,15,16,17,18,19,20,21,22,23},
                    xticklabels={0,1,2,3,4,5,10,15,20,25,30,35,40,45,50,55,60,65,70,75,80,85,90,95},
                    ytick={0.2,0.4,0.6,0.8,1},
                    legend pos=north west,
                    ymajorgrids=false,
                    xmajorgrids=false,
                    major grid style={line width=.1pt,draw=gray!50},
                    x axis line style={draw=black!60},
                    tick style={draw=black!60},
                ]

                \addplot[
                    color=cyan,
	                mark=none,
	                dashed,
	                thick
	               ]
	                coordinates { (1.0,0.203)(2.0,0.236)(3.0,0.365)(4.0,0.365)(5.0,0.426)(6.0,0.649)(7.0,0.736)(8.0,0.791)(9.0,0.858)(10.0,0.912)(11.0,0.946)(12.0,0.939)(13.0,0.953)(14.0,0.966)(15.0,0.973)(16.0,0.986)(17.0,0.993)(18.0,1.0)(19.0,1.0)(20.0,1.0)(21.0,1.0)(22.0,0.986)(23.0,0.966)};
                \addplot[
					color=green!70,
					mark=none,
					dashdotted,
					thick
					]
					coordinates { (1.0,0.394)(2.0,0.4)(3.0,0.423)(4.0,0.438)(5.0,0.45)(6.0,0.471)(7.0,0.479)(8.0,0.486)(9.0,0.491)(10.0,0.493)(11.0,0.494)(12.0,0.494)(13.0,0.494)(14.0,0.493)(15.0,0.494)(16.0,0.494)(17.0,0.495)(18.0,0.495)(19.0,0.495)(20.0,0.495)(21.0,0.495)(22.0,0.496)(23.0,0.496)};
                \addplot[
					color=red,
					mark=none,
					densely dotted,
					thick
					]
					coordinates { (1.0,0.391)(2.0,0.398)(3.0,0.467)(4.0,0.531)(5.0,0.546)(6.0,0.677)(7.0,0.784)(8.0,0.812)(9.0,0.88)(10.0,0.915)(11.0,0.946)(12.0,0.949)(13.0,0.953)(14.0,0.966)(15.0,0.973)(16.0,0.986)(17.0,0.993)(18.0,1.0)(19.0,1.0)(20.0,1.0)(21.0,1.0)(22.0,0.986)(23.0,0.966)};
				\addplot[
					color=black,
					mark=none,
					densely dotted,
					thick
					]
					coordinates { (1.0,0.317)(2.0,0.319)(3.0,0.437)(4.0,0.43)(5.0,0.507)(6.0,0.664)(7.0,0.777)(8.0,0.803)(9.0,0.88)(10.0,0.915)(11.0,0.946)(12.0,0.949)(13.0,0.953)(14.0,0.966)(15.0,0.973)(16.0,0.986)(17.0,0.993)(18.0,1.0)(19.0,1.0)(20.0,1.0)(21.0,1.0)(22.0,0.986)(23.0,0.966)};

                \end{axis}
            \end{tikzpicture}}
        \vspace{-0.25cm}
        \subfloat{
             \begin{tikzpicture}
            \pgfplotsset{every x tick label/.append style={font=\tiny}}
            \pgfplotsset{every y tick label/.append style={font=\tiny}}
                \begin{axis}[
                    width=0.55\linewidth,
                    height=4cm,
                    title={\footnotesize{n=5}},
                     title style={at={(0.063,0.915)},anchor=north,draw=black!60,fill=white},
                    ylabel={\scriptsize{$Acc(\lambda,CMU)$}},
                    xlabel near ticks,
                    ylabel near ticks,
                    xmin=0, xmax=24,
                    ymin=0, ymax=1.1,
                    xtick={0,1,2,3,4,5,6,7,8,9,10,11,12,13,14,15,16,17,18,19,20,21,22,23},
                    xticklabels={0,1,2,3,4,5,10,15,20,25,30,35,40,45,50,55,60,65,70,75,80,85,90,95},
                    ytick={0.2,0.4,0.6,0.8,1},
                    legend pos=north west,
                    ymajorgrids=false,
                    xmajorgrids=false,
                    major grid style={line width=.1pt,draw=gray!50},
                    x axis line style={draw=black!60},
                    tick style={draw=black!60},
                ]

                \addplot[
                    color=cyan,
	                mark=none,
	                dashed,
	                thick
	               ]
	                coordinates { (1.0,0.203)(2.0,0.236)(3.0,0.365)(4.0,0.365)(5.0,0.426)(6.0,0.649)(7.0,0.736)(8.0,0.791)(9.0,0.858)(10.0,0.912)(11.0,0.946)(12.0,0.939)(13.0,0.953)(14.0,0.966)(15.0,0.973)(16.0,0.986)(17.0,0.993)(18.0,1.0)(19.0,1.0)(20.0,1.0)(21.0,1.0)(22.0,0.986)(23.0,0.966)};
                \addplot[
					color=green!70,
					mark=none,
					dashdotted,
					thick
					]
					coordinates { (1.0,0.459)(2.0,0.467)(3.0,0.51)(4.0,0.53)(5.0,0.542)(6.0,0.569)(7.0,0.582)(8.0,0.591)(9.0,0.593)(10.0,0.6)(11.0,0.606)(12.0,0.604)(13.0,0.604)(14.0,0.606)(15.0,0.607)(16.0,0.609)(17.0,0.61)(18.0,0.612)(19.0,0.612)(20.0,0.611)(21.0,0.612)(22.0,0.611)(23.0,0.611)};
                \addplot[
					color=red,
					mark=none,
					densely dotted,
					thick
					]
					coordinates { (1.0,0.454)(2.0,0.466)(3.0,0.528)(4.0,0.57)(5.0,0.609)(6.0,0.708)(7.0,0.799)(8.0,0.83)(9.0,0.886)(10.0,0.92)(11.0,0.946)(12.0,0.951)(13.0,0.953)(14.0,0.966)(15.0,0.973)(16.0,0.986)(17.0,0.993)(18.0,1.0)(19.0,1.0)(20.0,1.0)(21.0,1.0)(22.0,0.986)(23.0,0.966)};
				\addplot[
					color=black,
					mark=none,
					densely dotted,
					thick
					]
					coordinates { (1.0,0.344)(2.0,0.336)(3.0,0.449)(4.0,0.444)(5.0,0.524)(6.0,0.667)(7.0,0.789)(8.0,0.806)(9.0,0.886)(10.0,0.916)(11.0,0.946)(12.0,0.951)(13.0,0.953)(14.0,0.966)(15.0,0.973)(16.0,0.986)(17.0,0.993)(18.0,1.0)(19.0,1.0)(20.0,1.0)(21.0,1.0)(22.0,0.986)(23.0,0.966)};

                \end{axis}
            \end{tikzpicture}}
        \subfloat{
             \begin{tikzpicture}
            \pgfplotsset{every x tick label/.append style={font=\tiny}}
            \pgfplotsset{every y tick label/.append style={font=\tiny}}
                \begin{axis}[
                    width=0.55\linewidth,
                    height=4cm,
                    title={\footnotesize{n=7}},
                    title style={at={(0.063,0.915)},anchor=north,draw=black!60,fill=white},
                    xlabel near ticks,
                    ylabel near ticks,
                    xmin=0, xmax=24,
                    ymin=0, ymax=1.1,
                    xtick={0,1,2,3,4,5,6,7,8,9,10,11,12,13,14,15,16,17,18,19,20,21,22,23},
                    xticklabels={0,1,2,3,4,5,10,15,20,25,30,35,40,45,50,55,60,65,70,75,80,85,90,95},
                    ytick={0.2,0.4,0.6,0.8,1},
                    legend pos=north west,
                    ymajorgrids=false,
                    xmajorgrids=false,
                    major grid style={line width=.1pt,draw=gray!50},
                    x axis line style={draw=black!60},
                    tick style={draw=black!60},
                ]

                \addplot[
                    color=cyan,
	                mark=none,
	                dashed,
	                thick
	               ]
	                coordinates { (1.0,0.203)(2.0,0.236)(3.0,0.365)(4.0,0.365)(5.0,0.426)(6.0,0.649)(7.0,0.736)(8.0,0.791)(9.0,0.858)(10.0,0.912)(11.0,0.946)(12.0,0.939)(13.0,0.953)(14.0,0.966)(15.0,0.973)(16.0,0.986)(17.0,0.993)(18.0,1.0)(19.0,1.0)(20.0,1.0)(21.0,1.0)(22.0,0.986)(23.0,0.966)};
                \addplot[
					color=green!70,
					mark=none,
					dashdotted,
					thick
					]
					coordinates { (1.0,0.51)(2.0,0.524)(3.0,0.589)(4.0,0.622)(5.0,0.642)(6.0,0.703)(7.0,0.726)(8.0,0.743)(9.0,0.751)(10.0,0.759)(11.0,0.764)(12.0,0.764)(13.0,0.764)(14.0,0.765)(15.0,0.765)(16.0,0.769)(17.0,0.771)(18.0,0.772)(19.0,0.771)(20.0,0.772)(21.0,0.773)(22.0,0.773)(23.0,0.772)};
                \addplot[
					color=red,
					mark=none,
					densely dotted,
					thick
					]
					coordinates { (1.0,0.504)(2.0,0.523)(3.0,0.614)(4.0,0.651)(5.0,0.68)(6.0,0.785)(7.0,0.851)(8.0,0.865)(9.0,0.886)(10.0,0.912)(11.0,0.94)(12.0,0.952)(13.0,0.953)(14.0,0.966)(15.0,0.973)(16.0,0.986)(17.0,0.993)(18.0,1.0)(19.0,1.0)(20.0,1.0)(21.0,1.0)(22.0,0.986)(23.0,0.966)};
				\addplot[
					color=black,
					mark=none,
					densely dotted,
					thick
					]
					coordinates { (1.0,0.362)(2.0,0.348)(3.0,0.465)(4.0,0.463)(5.0,0.55)(6.0,0.672)(7.0,0.801)(8.0,0.811)(9.0,0.893)(10.0,0.917)(11.0,0.946)(12.0,0.954)(13.0,0.953)(14.0,0.966)(15.0,0.973)(16.0,0.986)(17.0,0.993)(18.0,1.0)(19.0,1.0)(20.0,1.0)(21.0,1.0)(22.0,0.986)(23.0,0.966)};

                \end{axis}
            \end{tikzpicture}}
        \vspace{-0.25cm}
        \subfloat{
             \begin{tikzpicture}
            \pgfplotsset{every x tick label/.append style={font=\tiny}}
            \pgfplotsset{every y tick label/.append style={font=\tiny}}
                \begin{axis}[
                    width=0.55\linewidth,
                    height=4cm,
                    title={\footnotesize{n=9}},
                     title style={at={(0.063,0.915)},anchor=north,draw=black!60,fill=white},
                    xlabel={\scriptsize{$\lambda$ (\%)}},
                    ylabel={\scriptsize{$Acc(\lambda,CMU)$}},
                    xlabel near ticks,
                    ylabel near ticks,
                    xmin=0, xmax=24,
                    ymin=0, ymax=1.1,
                    xtick={0,1,2,3,4,5,6,7,8,9,10,11,12,13,14,15,16,17,18,19,20,21,22,23},
                    xticklabels={0,1,2,3,4,5,10,15,20,25,30,35,40,45,50,55,60,65,70,75,80,85,90,95},
                    ytick={0.2,0.4,0.6,0.8,1},
                    legend pos=north west,
                    ymajorgrids=false,
                    xmajorgrids=false,
                    major grid style={line width=.1pt,draw=gray!50},
                    x axis line style={draw=black!60},
                    tick style={draw=black!60},
                ]

                \addplot[
                    color=cyan,
	                mark=none,
	                dashed,
	                thick
	               ]
	                coordinates { (1.0,0.203)(2.0,0.236)(3.0,0.365)(4.0,0.365)(5.0,0.426)(6.0,0.649)(7.0,0.736)(8.0,0.791)(9.0,0.858)(10.0,0.912)(11.0,0.946)(12.0,0.939)(13.0,0.953)(14.0,0.966)(15.0,0.973)(16.0,0.986)(17.0,0.993)(18.0,1.0)(19.0,1.0)(20.0,1.0)(21.0,1.0)(22.0,0.986)(23.0,0.966)};
                \addplot[
					color=green!70,
					mark=none,
					dashdotted,
					thick
					]
					coordinates { (1.0,0.601)(2.0,0.614)(3.0,0.675)(4.0,0.718)(5.0,0.752)(6.0,0.807)(7.0,0.834)(8.0,0.85)(9.0,0.857)(10.0,0.865)(11.0,0.871)(12.0,0.871)(13.0,0.871)(14.0,0.874)(15.0,0.874)(16.0,0.876)(17.0,0.882)(18.0,0.883)(19.0,0.882)(20.0,0.882)(21.0,0.883)(22.0,0.884)(23.0,0.884)};
                \addplot[
					color=red,
					mark=none,
					densely dotted,
					thick
					]
					coordinates { (1.0,0.602)(2.0,0.612)(3.0,0.689)(4.0,0.74)(5.0,0.77)(6.0,0.849)(7.0,0.894)(8.0,0.895)(9.0,0.918)(10.0,0.93)(11.0,0.943)(12.0,0.955)(13.0,0.951)(14.0,0.966)(15.0,0.973)(16.0,0.986)(17.0,0.993)(18.0,1.0)(19.0,1.0)(20.0,1.0)(21.0,1.0)(22.0,0.986)(23.0,0.966)};
				\addplot[
					color=black,
					mark=none,
					densely dotted,
					thick
					]
					coordinates { (1.0,0.386)(2.0,0.365)(3.0,0.478)(4.0,0.479)(5.0,0.569)(6.0,0.677)(7.0,0.816)(8.0,0.815)(9.0,0.901)(10.0,0.918)(11.0,0.946)(12.0,0.958)(13.0,0.953)(14.0,0.966)(15.0,0.973)(16.0,0.986)(17.0,0.993)(18.0,1.0)(19.0,1.0)(20.0,1.0)(21.0,1.0)(22.0,0.986)(23.0,0.966)};

                \end{axis}
            \end{tikzpicture}}
        \subfloat{
             \begin{tikzpicture}
            \pgfplotsset{every x tick label/.append style={font=\tiny}}
            \pgfplotsset{every y tick label/.append style={font=\tiny}}
                \begin{axis}[
                    width=0.55\linewidth,
                    height=4cm,
                    title={\footnotesize{n=11}},
                    title style={at={(0.0745,0.915)},anchor=north,draw=black!60,fill=white},
                    xlabel={\scriptsize{$\lambda$ (\%)}},
                    xlabel near ticks,
                    ylabel near ticks,
                    xmin=0, xmax=24,
                    ymin=0, ymax=1.1,
                    xtick={0,1,2,3,4,5,6,7,8,9,10,11,12,13,14,15,16,17,18,19,20,21,22,23},
                    xticklabels={0,1,2,3,4,5,10,15,20,25,30,35,40,45,50,55,60,65,70,75,80,85,90,95},
                    ytick={0.2,0.4,0.6,0.8,1},
                    legend pos=north west,
                    ymajorgrids=false,
                    xmajorgrids=false,
                    major grid style={line width=.1pt,draw=gray!50},
                    x axis line style={draw=black!60},
                    tick style={draw=black!60},
                ]

                \addplot[
                    color=cyan,
	                mark=none,
	                dashed,
	                thick
	               ]
	                coordinates { (1.0,0.203)(2.0,0.236)(3.0,0.365)(4.0,0.365)(5.0,0.426)(6.0,0.649)(7.0,0.736)(8.0,0.791)(9.0,0.858)(10.0,0.912)(11.0,0.946)(12.0,0.939)(13.0,0.953)(14.0,0.966)(15.0,0.973)(16.0,0.986)(17.0,0.993)(18.0,1.0)(19.0,1.0)(20.0,1.0)(21.0,1.0)(22.0,0.986)(23.0,0.966)};
                \addplot[
					color=green!70,
					mark=none,
					dashdotted,
					thick
					]
					coordinates { (1.0,0.648)(2.0,0.657)(3.0,0.726)(4.0,0.755)(5.0,0.778)(6.0,0.829)(7.0,0.848)(8.0,0.855)(9.0,0.868)(10.0,0.876)(11.0,0.886)(12.0,0.884)(13.0,0.886)(14.0,0.885)(15.0,0.885)(16.0,0.886)(17.0,0.89)(18.0,0.889)(19.0,0.89)(20.0,0.889)(21.0,0.89)(22.0,0.893)(23.0,0.892)};
                \addplot[
					color=red,
					mark=none,
					densely dotted,
					thick
					]
					coordinates { (1.0,0.653)(2.0,0.666)(3.0,0.755)(4.0,0.769)(5.0,0.785)(6.0,0.857)(7.0,0.898)(8.0,0.884)(9.0,0.91)(10.0,0.929)(11.0,0.936)(12.0,0.949)(13.0,0.95)(14.0,0.966)(15.0,0.973)(16.0,0.986)(17.0,0.993)(18.0,1.0)(19.0,1.0)(20.0,1.0)(21.0,1.0)(22.0,0.986)(23.0,0.966)};
				\addplot[
					color=black,
					mark=none,
					densely dotted,
					thick
					]
					coordinates { (1.0,0.405)(2.0,0.379)(3.0,0.489)(4.0,0.483)(5.0,0.575)(6.0,0.675)(7.0,0.813)(8.0,0.814)(9.0,0.899)(10.0,0.918)(11.0,0.946)(12.0,0.957)(13.0,0.953)(14.0,0.966)(15.0,0.973)(16.0,0.986)(17.0,0.993)(18.0,1.0)(19.0,1.0)(20.0,1.0)(21.0,1.0)(22.0,0.986)(23.0,0.966)};

                \end{axis}
            \end{tikzpicture}}
            \vspace{-0.1cm}
    \ref{plotLabel2}\vspace{-0.3cm}
    \caption{Mean accuracy of the Goal Mirroring (GM) and Naive Bayes Model (NBM) approaches and the two hybrid approaches Weighted Sum (WS) and Tiebreaking (TB) on the \emph{CMU Dataset without artifical samples} for different sizes of the training set $n$.}
    \label{fig:CMUHybridEvaluation}
\end{figure*}
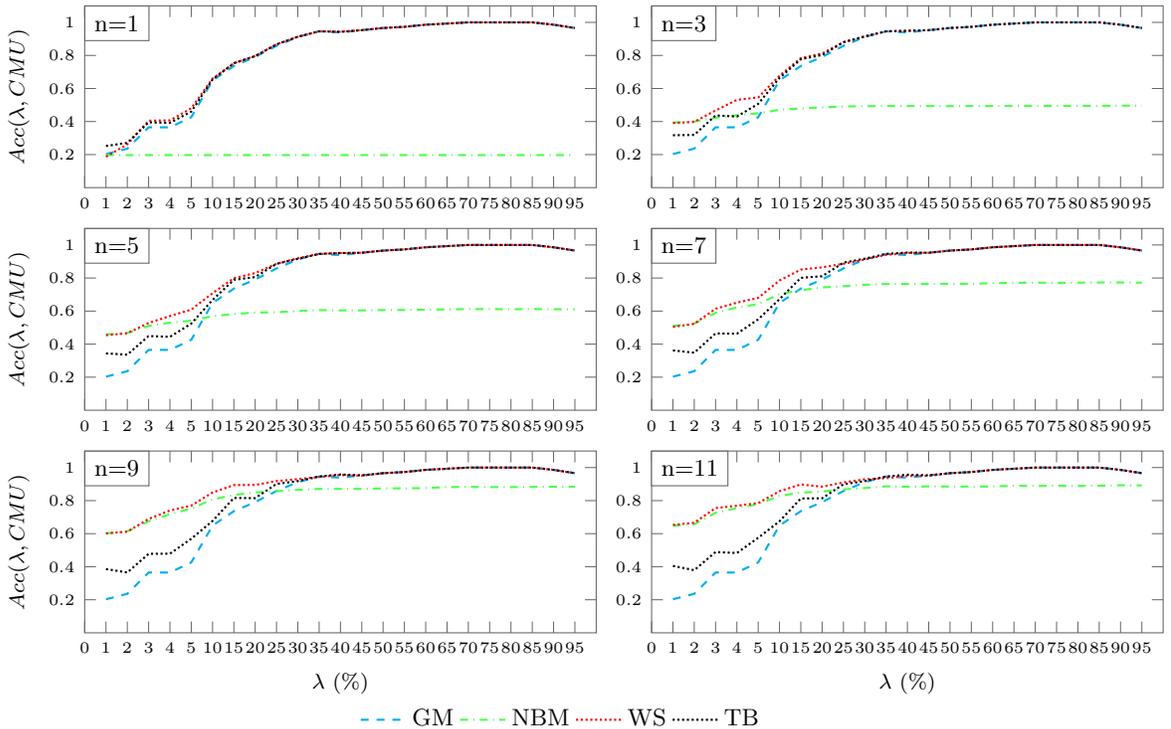

%% file: figures/figures_tex/evaluationFigure_CMU_artificial.tex
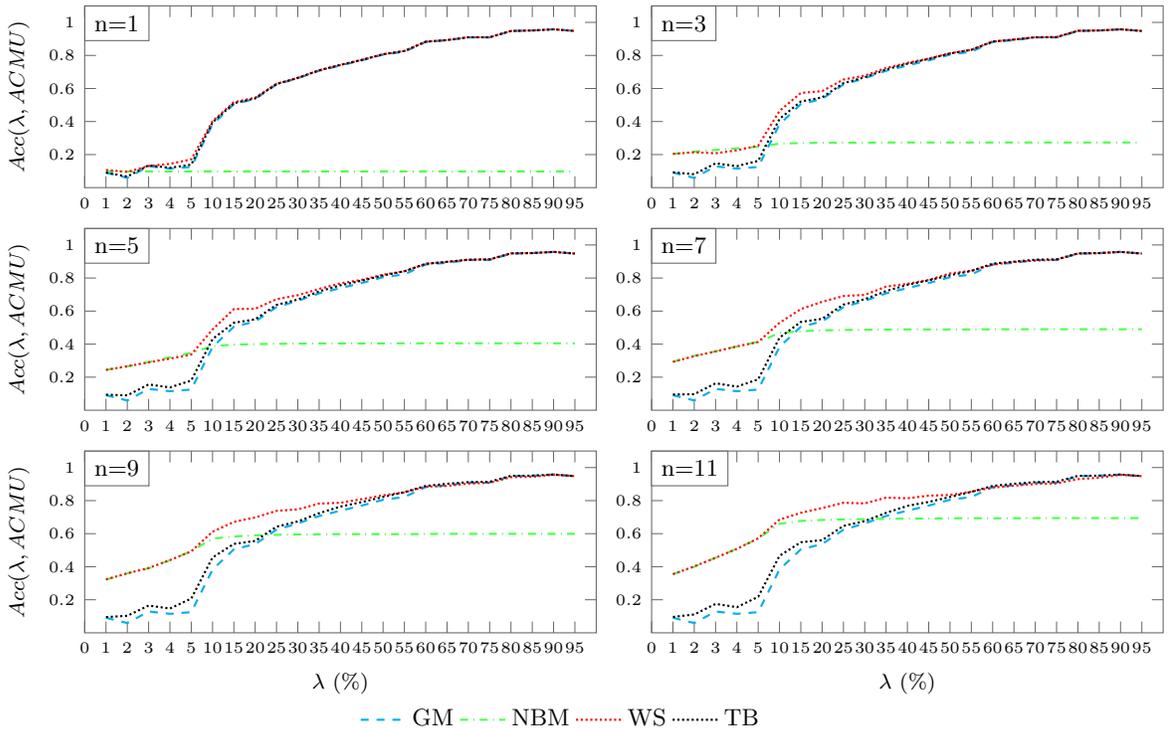
\begin{figure*}[t]
    \centering
        \subfloat{
            \begin{tikzpicture}
            \pgfplotsset{every x tick label/.append style={font=\tiny}}
            \pgfplotsset{every y tick label/.append style={font=\tiny}}
                \begin{axis}[
                    width=0.55\linewidth,
                    height=4cm,
                    title={\footnotesize{n=1}},
                    title style={at={(0.063,0.915)},anchor=north,draw=black!60,fill=white},
                    ylabel={\scriptsize{$Acc(\lambda,ACMU)$}},
                    xlabel near ticks,
                    ylabel near ticks,
                    xmin=0, xmax=24,
                    ymin=0, ymax=1.1,
                    xtick={0,1,2,3,4,5,6,7,8,9,10,11,12,13,14,15,16,17,18,19,20,21,22,23},
                    xticklabels={0,1,2,3,4,5,10,15,20,25,30,35,40,45,50,55,60,65,70,75,80,85,90,95},
                    ytick={0.2,0.4,0.6,0.8,1},
                    legend pos=north west,
                    ymajorgrids=false,
                    xmajorgrids=false,
                    major grid style={line width=.1pt,draw=gray!50},
                    x axis line style={draw=black!60},
                    tick style={draw=black!60},
                    legend columns=4,
                    legend style={draw=none},
                    legend entries={\footnotesize{GM},\footnotesize{NBM},\footnotesize{WS},\footnotesize{TB}},
                    legend to name={plotLabel1}
                ]
	           \addplot[
                    color=cyan,
	                mark=none,
	                dashed,
	                thick
	               ]
	                coordinates { (1.0,0.091)(2.0,0.059)(3.0,0.129)(4.0,0.115)(5.0,0.125)(6.0,0.383)(7.0,0.505)(8.0,0.537)(9.0,0.624)(10.0,0.662)(11.0,0.707)(12.0,0.739)(13.0,0.77)(14.0,0.805)(15.0,0.822)(16.0,0.882)(17.0,0.892)(18.0,0.909)(19.0,0.909)(20.0,0.948)(21.0,0.951)(22.0,0.958)(23.0,0.948)};
                \addplot[
					color=green!70,
					mark=none,
					dashdotted,
					thick
					]
					coordinates { (1.0,0.098)(2.0,0.098)(3.0,0.098)(4.0,0.098)(5.0,0.098)(6.0,0.098)(7.0,0.098)(8.0,0.098)(9.0,0.098)(10.0,0.098)(11.0,0.098)(12.0,0.098)(13.0,0.098)(14.0,0.098)(15.0,0.098)(16.0,0.098)(17.0,0.098)(18.0,0.098)(19.0,0.098)(20.0,0.098)(21.0,0.098)(22.0,0.098)(23.0,0.098)};
                \addplot[
					color=red,
					mark=none,
					densely dotted,
					thick
					]
					coordinates { (1.0,0.107)(2.0,0.095)(3.0,0.131)(4.0,0.144)(5.0,0.171)(6.0,0.402)(7.0,0.517)(8.0,0.544)(9.0,0.628)(10.0,0.664)(11.0,0.71)(12.0,0.742)(13.0,0.773)(14.0,0.807)(15.0,0.827)(16.0,0.883)(17.0,0.893)(18.0,0.91)(19.0,0.91)(20.0,0.948)(21.0,0.951)(22.0,0.958)(23.0,0.948)};
				\addplot[
					color=black,
					mark=none,
					densely dotted,
					thick
					]
					coordinates { (1.0,0.091)(2.0,0.067)(3.0,0.135)(4.0,0.12)(5.0,0.139)(6.0,0.395)(7.0,0.511)(8.0,0.54)(9.0,0.627)(10.0,0.664)(11.0,0.71)(12.0,0.743)(13.0,0.773)(14.0,0.807)(15.0,0.827)(16.0,0.883)(17.0,0.893)(18.0,0.91)(19.0,0.91)(20.0,0.948)(21.0,0.951)(22.0,0.958)(23.0,0.948)};

                \end{axis}
            \end{tikzpicture}}
        \subfloat{
             \begin{tikzpicture}
            \pgfplotsset{every x tick label/.append style={font=\tiny}}
            \pgfplotsset{every y tick label/.append style={font=\tiny}}
                \begin{axis}[
                    width=0.55\linewidth,
                    height=4cm,
                    title={\footnotesize{n=3}},
                     title style={at={(0.063,0.915)},anchor=north,draw=black!60,fill=white},
                    xlabel near ticks,
                    ylabel near ticks,
                    xmin=0, xmax=24,
                    ymin=0, ymax=1.1,
                    xtick={0,1,2,3,4,5,6,7,8,9,10,11,12,13,14,15,16,17,18,19,20,21,22,23},
                    xticklabels={0,1,2,3,4,5,10,15,20,25,30,35,40,45,50,55,60,65,70,75,80,85,90,95},
                    ytick={0.2,0.4,0.6,0.8,1},
                    legend pos=north west,
                    ymajorgrids=false,
                    xmajorgrids=false,
                    major grid style={line width=.1pt,draw=gray!50},
                    x axis line style={draw=black!60},
                    tick style={draw=black!60},
                ]
	            \addplot[
                    color=cyan,
	                mark=none,
	                dashed,
	                thick
	               ]
	                coordinates { (1.0,0.091)(2.0,0.059)(3.0,0.129)(4.0,0.115)(5.0,0.125)(6.0,0.383)(7.0,0.505)(8.0,0.537)(9.0,0.624)(10.0,0.662)(11.0,0.707)(12.0,0.739)(13.0,0.77)(14.0,0.805)(15.0,0.822)(16.0,0.882)(17.0,0.892)(18.0,0.909)(19.0,0.909)(20.0,0.948)(21.0,0.951)(22.0,0.958)(23.0,0.948)};
                \addplot[
					color=green!70,
					mark=none,
					dashdotted,
					thick
					]
					coordinates { (1.0,0.204)(2.0,0.218)(3.0,0.228)(4.0,0.238)(5.0,0.247)(6.0,0.266)(7.0,0.27)(8.0,0.271)(9.0,0.272)(10.0,0.272)(11.0,0.273)(12.0,0.273)(13.0,0.273)(14.0,0.273)(15.0,0.273)(16.0,0.273)(17.0,0.273)(18.0,0.273)(19.0,0.273)(20.0,0.273)(21.0,0.273)(22.0,0.273)(23.0,0.273)};
                \addplot[
					color=red,
					mark=none,
					densely dotted,
					thick
					]
					coordinates { (1.0,0.204)(2.0,0.214)(3.0,0.208)(4.0,0.226)(5.0,0.254)(6.0,0.464)(7.0,0.573)(8.0,0.584)(9.0,0.654)(10.0,0.678)(11.0,0.724)(12.0,0.755)(13.0,0.781)(14.0,0.813)(15.0,0.834)(16.0,0.884)(17.0,0.896)(18.0,0.91)(19.0,0.911)(20.0,0.949)(21.0,0.951)(22.0,0.958)(23.0,0.948)};
				\addplot[
					color=black,
					mark=none,
					densely dotted,
					thick
					]
					coordinates { (1.0,0.093)(2.0,0.081)(3.0,0.147)(4.0,0.13)(5.0,0.162)(6.0,0.416)(7.0,0.521)(8.0,0.546)(9.0,0.632)(10.0,0.668)(11.0,0.715)(12.0,0.75)(13.0,0.779)(14.0,0.812)(15.0,0.834)(16.0,0.884)(17.0,0.896)(18.0,0.91)(19.0,0.911)(20.0,0.949)(21.0,0.951)(22.0,0.958)(23.0,0.948)};

                \end{axis}
            \end{tikzpicture}}
        \vspace{-0.25cm}
        \subfloat{
             \begin{tikzpicture}
            \pgfplotsset{every x tick label/.append style={font=\tiny}}
            \pgfplotsset{every y tick label/.append style={font=\tiny}}
                \begin{axis}[
                    width=0.55\linewidth,
                    height=4cm,
                    title={\footnotesize{n=5}},
                     title style={at={(0.063,0.915)},anchor=north,draw=black!60,fill=white},
                    ylabel={\scriptsize{$Acc(\lambda,ACMU)$}},
                    xlabel near ticks,
                    ylabel near ticks,
                    xmin=0, xmax=24,
                    ymin=0, ymax=1.1,
                    xtick={0,1,2,3,4,5,6,7,8,9,10,11,12,13,14,15,16,17,18,19,20,21,22,23},
                    xticklabels={0,1,2,3,4,5,10,15,20,25,30,35,40,45,50,55,60,65,70,75,80,85,90,95},
                    ytick={0.2,0.4,0.6,0.8,1},
                    legend pos=north west,
                    ymajorgrids=false,
                    xmajorgrids=false,
                    major grid style={line width=.1pt,draw=gray!50},
                    x axis line style={draw=black!60},
                    tick style={draw=black!60},
                ]
	            \addplot[
                    color=cyan,
	                mark=none,
	                dashed,
	                thick
	               ]
	                coordinates { (1.0,0.091)(2.0,0.059)(3.0,0.129)(4.0,0.115)(5.0,0.125)(6.0,0.383)(7.0,0.505)(8.0,0.537)(9.0,0.624)(10.0,0.662)(11.0,0.707)(12.0,0.739)(13.0,0.77)(14.0,0.805)(15.0,0.822)(16.0,0.882)(17.0,0.892)(18.0,0.909)(19.0,0.909)(20.0,0.948)(21.0,0.951)(22.0,0.958)(23.0,0.948)};
                \addplot[
					color=green!70,
					mark=none,
					dashdotted,
					thick
					]
					coordinates { (1.0,0.244)(2.0,0.265)(3.0,0.291)(4.0,0.32)(5.0,0.348)(6.0,0.388)(7.0,0.397)(8.0,0.4)(9.0,0.402)(10.0,0.403)(11.0,0.404)(12.0,0.404)(13.0,0.404)(14.0,0.404)(15.0,0.405)(16.0,0.405)(17.0,0.405)(18.0,0.405)(19.0,0.405)(20.0,0.405)(21.0,0.405)(22.0,0.405)(23.0,0.405)};
                \addplot[
					color=red,
					mark=none,
					densely dotted,
					thick
					]
					coordinates { (1.0,0.244)(2.0,0.266)(3.0,0.29)(4.0,0.312)(5.0,0.337)(6.0,0.489)(7.0,0.612)(8.0,0.614)(9.0,0.672)(10.0,0.696)(11.0,0.734)(12.0,0.768)(13.0,0.789)(14.0,0.82)(15.0,0.841)(16.0,0.886)(17.0,0.898)(18.0,0.91)(19.0,0.913)(20.0,0.949)(21.0,0.951)(22.0,0.958)(23.0,0.948)};
				\addplot[
					color=black,
					mark=none,
					densely dotted,
					thick
					]
					coordinates { (1.0,0.094)(2.0,0.09)(3.0,0.156)(4.0,0.138)(5.0,0.18)(6.0,0.431)(7.0,0.529)(8.0,0.55)(9.0,0.637)(10.0,0.67)(11.0,0.719)(12.0,0.756)(13.0,0.784)(14.0,0.816)(15.0,0.841)(16.0,0.886)(17.0,0.898)(18.0,0.911)(19.0,0.913)(20.0,0.949)(21.0,0.951)(22.0,0.958)(23.0,0.948)};

                \end{axis}
            \end{tikzpicture}}
        \subfloat{
             \begin{tikzpicture}
            \pgfplotsset{every x tick label/.append style={font=\tiny}}
            \pgfplotsset{every y tick label/.append style={font=\tiny}}
                \begin{axis}[
                    width=0.55\linewidth,
                    height=4cm,
                    title={\footnotesize{n=7}},
                    title style={at={(0.063,0.915)},anchor=north,draw=black!60,fill=white},
                    xlabel near ticks,
                    ylabel near ticks,
                    xmin=0, xmax=24,
                    ymin=0, ymax=1.1,
                    xtick={0,1,2,3,4,5,6,7,8,9,10,11,12,13,14,15,16,17,18,19,20,21,22,23},
                    xticklabels={0,1,2,3,4,5,10,15,20,25,30,35,40,45,50,55,60,65,70,75,80,85,90,95},
                    ytick={0.2,0.4,0.6,0.8,1},
                    legend pos=north west,
                    ymajorgrids=false,
                    xmajorgrids=false,
                    major grid style={line width=.1pt,draw=gray!50},
                    x axis line style={draw=black!60},
                    tick style={draw=black!60},
                ]
	            \addplot[
                    color=cyan,
	                mark=none,
	                dashed,
	                thick
	               ]
	                coordinates { (1.0,0.091)(2.0,0.059)(3.0,0.129)(4.0,0.115)(5.0,0.125)(6.0,0.383)(7.0,0.505)(8.0,0.537)(9.0,0.624)(10.0,0.662)(11.0,0.707)(12.0,0.739)(13.0,0.77)(14.0,0.805)(15.0,0.822)(16.0,0.882)(17.0,0.892)(18.0,0.909)(19.0,0.909)(20.0,0.948)(21.0,0.951)(22.0,0.958)(23.0,0.948)};
                \addplot[
					color=green!70,
					mark=none,
					dashdotted,
					thick
					]
					coordinates { (1.0,0.293)(2.0,0.329)(3.0,0.356)(4.0,0.386)(5.0,0.414)(6.0,0.467)(7.0,0.478)(8.0,0.483)(9.0,0.486)(10.0,0.487)(11.0,0.488)(12.0,0.489)(13.0,0.489)(14.0,0.489)(15.0,0.489)(16.0,0.49)(17.0,0.49)(18.0,0.49)(19.0,0.49)(20.0,0.49)(21.0,0.49)(22.0,0.49)(23.0,0.49)};
                \addplot[
					color=red,
					mark=none,
					densely dotted,
					thick
					]
					coordinates { (1.0,0.293)(2.0,0.327)(3.0,0.356)(4.0,0.386)(5.0,0.413)(6.0,0.529)(7.0,0.612)(8.0,0.656)(9.0,0.691)(10.0,0.698)(11.0,0.748)(12.0,0.765)(13.0,0.788)(14.0,0.828)(15.0,0.844)(16.0,0.881)(17.0,0.897)(18.0,0.905)(19.0,0.911)(20.0,0.948)(21.0,0.95)(22.0,0.958)(23.0,0.948)};
				\addplot[
					color=black,
					mark=none,
					densely dotted,
					thick
					]
					coordinates { (1.0,0.095)(2.0,0.097)(3.0,0.163)(4.0,0.143)(5.0,0.187)(6.0,0.439)(7.0,0.534)(8.0,0.553)(9.0,0.639)(10.0,0.672)(11.0,0.721)(12.0,0.759)(13.0,0.786)(14.0,0.817)(15.0,0.843)(16.0,0.887)(17.0,0.899)(18.0,0.911)(19.0,0.913)(20.0,0.949)(21.0,0.951)(22.0,0.958)(23.0,0.948)};

                \end{axis}
            \end{tikzpicture}}
        \vspace{-0.25cm}
        \subfloat{
             \begin{tikzpicture}
            \pgfplotsset{every x tick label/.append style={font=\tiny}}
            \pgfplotsset{every y tick label/.append style={font=\tiny}}
                \begin{axis}[
                    width=0.55\linewidth,
                    height=4cm,
                    title={\footnotesize{n=9}},
                     title style={at={(0.063,0.915)},anchor=north,draw=black!60,fill=white},
                    xlabel={\scriptsize{$\lambda$ (\%)}},
                    ylabel={\scriptsize{$Acc(\lambda,ACMU)$}},
                    xlabel near ticks,
                    ylabel near ticks,
                    xmin=0, xmax=24,
                    ymin=0, ymax=1.1,
                    xtick={0,1,2,3,4,5,6,7,8,9,10,11,12,13,14,15,16,17,18,19,20,21,22,23},
                    xticklabels={0,1,2,3,4,5,10,15,20,25,30,35,40,45,50,55,60,65,70,75,80,85,90,95},
                    ytick={0.2,0.4,0.6,0.8,1},
                    legend pos=north west,
                    ymajorgrids=false,
                    xmajorgrids=false,
                    major grid style={line width=.1pt,draw=gray!50},
                    x axis line style={draw=black!60},
                    tick style={draw=black!60},
                ]
	            \addplot[
                    color=cyan,
	                mark=none,
	                dashed,
	                thick
	               ]
	                coordinates { (1.0,0.091)(2.0,0.059)(3.0,0.129)(4.0,0.115)(5.0,0.125)(6.0,0.383)(7.0,0.505)(8.0,0.537)(9.0,0.624)(10.0,0.662)(11.0,0.707)(12.0,0.739)(13.0,0.77)(14.0,0.805)(15.0,0.822)(16.0,0.882)(17.0,0.892)(18.0,0.909)(19.0,0.909)(20.0,0.948)(21.0,0.951)(22.0,0.958)(23.0,0.948)};
                \addplot[
					color=green!70,
					mark=none,
					dashdotted,
					thick
					]
					coordinates { (1.0,0.322)(2.0,0.36)(3.0,0.39)(4.0,0.44)(5.0,0.494)(6.0,0.57)(7.0,0.583)(8.0,0.59)(9.0,0.593)(10.0,0.595)(11.0,0.596)(12.0,0.597)(13.0,0.597)(14.0,0.597)(15.0,0.597)(16.0,0.598)(17.0,0.599)(18.0,0.599)(19.0,0.599)(20.0,0.599)(21.0,0.599)(22.0,0.599)(23.0,0.599)};
                \addplot[
					color=red,
					mark=none,
					densely dotted,
					thick
					]
					coordinates { (1.0,0.323)(2.0,0.36)(3.0,0.392)(4.0,0.44)(5.0,0.493)(6.0,0.612)(7.0,0.671)(8.0,0.698)(9.0,0.738)(10.0,0.747)(11.0,0.782)(12.0,0.787)(13.0,0.809)(14.0,0.832)(15.0,0.85)(16.0,0.884)(17.0,0.889)(18.0,0.903)(19.0,0.908)(20.0,0.942)(21.0,0.944)(22.0,0.957)(23.0,0.948)};
				\addplot[
					color=black,
					mark=none,
					densely dotted,
					thick
					]
					coordinates { (1.0,0.095)(2.0,0.103)(3.0,0.165)(4.0,0.147)(5.0,0.207)(6.0,0.456)(7.0,0.538)(8.0,0.556)(9.0,0.642)(10.0,0.674)(11.0,0.724)(12.0,0.764)(13.0,0.791)(14.0,0.822)(15.0,0.851)(16.0,0.889)(17.0,0.902)(18.0,0.912)(19.0,0.914)(20.0,0.95)(21.0,0.951)(22.0,0.958)(23.0,0.948)};

                \end{axis}
            \end{tikzpicture}}
        \subfloat{
             \begin{tikzpicture}
            \pgfplotsset{every x tick label/.append style={font=\tiny}}
            \pgfplotsset{every y tick label/.append style={font=\tiny}}
                \begin{axis}[
                    width=0.55\linewidth,
                    height=4cm,
                    title={\footnotesize{n=11}},
                    title style={at={(0.0745,0.915)},anchor=north,draw=black!60,fill=white},
                    xlabel={\scriptsize{$\lambda$ (\%)}},
                    xlabel near ticks,
                    ylabel near ticks,
                    xmin=0, xmax=24,
                    ymin=0, ymax=1.1,
                    xtick={0,1,2,3,4,5,6,7,8,9,10,11,12,13,14,15,16,17,18,19,20,21,22,23},
                    xticklabels={0,1,2,3,4,5,10,15,20,25,30,35,40,45,50,55,60,65,70,75,80,85,90,95},
                    ytick={0.2,0.4,0.6,0.8,1},
                    legend pos=north west,
                    ymajorgrids=false,
                    xmajorgrids=false,
                    major grid style={line width=.1pt,draw=gray!50},
                    x axis line style={draw=black!60},
                    tick style={draw=black!60},
                ]
	            \addplot[
                    color=cyan,
	                mark=none,
	                dashed,
	                thick
	               ]
	                coordinates { (1.0,0.091)(2.0,0.059)(3.0,0.129)(4.0,0.115)(5.0,0.125)(6.0,0.383)(7.0,0.505)(8.0,0.537)(9.0,0.624)(10.0,0.662)(11.0,0.707)(12.0,0.739)(13.0,0.77)(14.0,0.805)(15.0,0.822)(16.0,0.882)(17.0,0.892)(18.0,0.909)(19.0,0.909)(20.0,0.948)(21.0,0.951)(22.0,0.958)(23.0,0.948)};
                \addplot[
					color=green!70,
					mark=none,
					dashdotted,
					thick
					]
					coordinates { (1.0,0.355)(2.0,0.4)(3.0,0.454)(4.0,0.509)(5.0,0.57)(6.0,0.66)(7.0,0.677)(8.0,0.683)(9.0,0.686)(10.0,0.688)(11.0,0.69)(12.0,0.691)(13.0,0.691)(14.0,0.693)(15.0,0.693)(16.0,0.693)(17.0,0.693)(18.0,0.694)(19.0,0.694)(20.0,0.694)(21.0,0.694)(22.0,0.694)(23.0,0.694)};
                \addplot[
					color=red,
					mark=none,
					densely dotted,
					thick
					]
					coordinates { (1.0,0.355)(2.0,0.402)(3.0,0.454)(4.0,0.508)(5.0,0.572)(6.0,0.685)(7.0,0.726)(8.0,0.755)(9.0,0.787)(10.0,0.783)(11.0,0.818)(12.0,0.814)(13.0,0.83)(14.0,0.834)(15.0,0.854)(16.0,0.879)(17.0,0.89)(18.0,0.901)(19.0,0.904)(20.0,0.93)(21.0,0.939)(22.0,0.956)(23.0,0.948)};
				\addplot[
					color=black,
					mark=none,
					densely dotted,
					thick
					]
					coordinates { (1.0,0.096)(2.0,0.111)(3.0,0.175)(4.0,0.155)(5.0,0.218)(6.0,0.466)(7.0,0.548)(8.0,0.561)(9.0,0.647)(10.0,0.676)(11.0,0.726)(12.0,0.768)(13.0,0.792)(14.0,0.822)(15.0,0.853)(16.0,0.889)(17.0,0.902)(18.0,0.912)(19.0,0.914)(20.0,0.95)(21.0,0.951)(22.0,0.958)(23.0,0.948)};

                \end{axis}
            \end{tikzpicture}}
            \vspace{-0.1cm}
    \ref{plotLabel1}\vspace{-0.3cm}
    \caption{Mean accuracy of the Goal Mirroring (GM), Naive Bayes Model (NBM) approaches and the two hybrid approaches Weighted Sum (WS) and Tiebreaking (TB) on the \emph{artificially extended CMU Dataset} for different sizes of the training set $n$. }
    \label{fig:CMUArtificialHybridEvaluation}
\end{figure*}

%% file: figures/figures_tex/evaluationFigureLogistics.tex
\begin{figure*}[t]
    \centering
        \subfloat{
            \begin{tikzpicture}
            \pgfplotsset{every x tick label/.append style={font=\tiny}}
            \pgfplotsset{every y tick label/.append style={font=\tiny}}
                \begin{axis}[
                    width=0.55\linewidth,
                    height=4cm,
                    title={\footnotesize{n=1}},
                    title style={at={(0.063,0.915)},anchor=north,draw=black!60,fill=white},
                    ylabel={\scriptsize{$Acc(\lambda,LOG)$}},
                    xlabel near ticks,
                    ylabel near ticks,
                    xmin=0, xmax=24,
                    ymin=0, ymax=1.1,
                    xtick={0,1,2,3,4,5,6,7,8,9,10,11,12,13,14,15,16,17,18,19,20,21,22,23},
                    xticklabels={0,1,2,3,4,5,10,15,20,25,30,35,40,45,50,55,60,65,70,75,80,85,90,95},
                    ytick={0.2,0.4,0.6,0.8,1},
                    legend pos=north west,
                    ymajorgrids=false,
                    xmajorgrids=false,
                    major grid style={line width=.1pt,draw=gray!50},
                    x axis line style={draw=black!60},
                    tick style={draw=black!60},
                    legend columns=4,
                    legend style={draw=none},
                    legend entries={\footnotesize{GM},\footnotesize{NBM},\footnotesize{WS},\footnotesize{TB}},
                    legend to name={plotLabel5}
                ]
                \addplot[
                    color=cyan,
	                mark=none,
	                thick,
	                dashed
	               ]
	                coordinates { (1.0,0.0)(2.0,0.0)(3.0,0.0)(4.0,0.077)(5.0,0.21)(6.0,0.303)(7.0,0.383)(8.0,0.537)(9.0,0.597)(10.0,0.657)(11.0,0.693)(12.0,0.74)(13.0,0.77)(14.0,0.793)(15.0,0.853)(16.0,0.883)(17.0,0.883)(18.0,0.897)(19.0,0.897)(20.0,0.897)(21.0,0.897)(22.0,0.917)(23.0,0.99)};
                \addplot[
					color=green!70,
					mark=none,
					thick,
					dashdotted
					]
					coordinates { (1.0,0.097)(2.0,0.097)(3.0,0.097)(4.0,0.097)(5.0,0.097)(6.0,0.097)(7.0,0.097)(8.0,0.097)(9.0,0.097)(10.0,0.097)(11.0,0.097)(12.0,0.097)(13.0,0.097)(14.0,0.097)(15.0,0.097)(16.0,0.097)(17.0,0.097)(18.0,0.097)(19.0,0.097)(20.0,0.097)(21.0,0.097)(22.0,0.097)(23.0,0.097)};
                \addplot[
					color=red,
					mark=none,
					thick,
					densely dotted
					]
					coordinates { (1.0,0.097)(2.0,0.097)(3.0,0.097)(4.0,0.165)(5.0,0.279)(6.0,0.358)(7.0,0.432)(8.0,0.57)(9.0,0.623)(10.0,0.674)(11.0,0.704)(12.0,0.746)(13.0,0.775)(14.0,0.802)(15.0,0.857)(16.0,0.885)(17.0,0.885)(18.0,0.897)(19.0,0.899)(20.0,0.904)(21.0,0.906)(22.0,0.925)(23.0,0.991)};
				\addplot[
					color=black,
					mark=none,
					thick,
					densely dotted
					]
					coordinates { (1.0,0.078)(2.0,0.078)(3.0,0.078)(4.0,0.152)(5.0,0.274)(6.0,0.358)(7.0,0.432)(8.0,0.57)(9.0,0.623)(10.0,0.674)(11.0,0.704)(12.0,0.746)(13.0,0.775)(14.0,0.802)(15.0,0.857)(16.0,0.885)(17.0,0.885)(18.0,0.897)(19.0,0.899)(20.0,0.904)(21.0,0.906)(22.0,0.925)(23.0,0.991)};
                \end{axis}
            \end{tikzpicture}}
        \subfloat{
             \begin{tikzpicture}
            \pgfplotsset{every x tick label/.append style={font=\tiny}}
            \pgfplotsset{every y tick label/.append style={font=\tiny}}
                \begin{axis}[
                    width=0.55\linewidth,
                    height=4cm,
                    title={\footnotesize{n=3}},
                     title style={at={(0.063,0.915)},anchor=north,draw=black!60,fill=white},
                    xlabel near ticks,
                    ylabel near ticks,
                    xmin=0, xmax=24,
                    ymin=0, ymax=1.1,
                    xtick={0,1,2,3,4,5,6,7,8,9,10,11,12,13,14,15,16,17,18,19,20,21,22,23},
                    xticklabels={0,1,2,3,4,5,10,15,20,25,30,35,40,45,50,55,60,65,70,75,80,85,90,95},
                    ytick={0.2,0.4,0.6,0.8,1},
                    legend pos=north west,
                    ymajorgrids=false,
                    xmajorgrids=false,
                    major grid style={line width=.1pt,draw=gray!50},
                    x axis line style={draw=black!60},
                    tick style={draw=black!60},
                ]
                \addplot[
                    color=cyan,
	                mark=none,
	                thick,
	                dashed
	               ]
	                coordinates { (1.0,0.0)(2.0,0.0)(3.0,0.0)(4.0,0.077)(5.0,0.21)(6.0,0.303)(7.0,0.383)(8.0,0.537)(9.0,0.597)(10.0,0.657)(11.0,0.693)(12.0,0.74)(13.0,0.77)(14.0,0.793)(15.0,0.853)(16.0,0.883)(17.0,0.883)(18.0,0.897)(19.0,0.897)(20.0,0.897)(21.0,0.897)(22.0,0.917)(23.0,0.99)};
                \addplot[
					color=green!70,
					mark=none,
					thick,
					dashdotted
					]
					coordinates { (1.0,0.097)(2.0,0.097)(3.0,0.097)(4.0,0.118)(5.0,0.165)(6.0,0.201)(7.0,0.227)(8.0,0.238)(9.0,0.243)(10.0,0.247)(11.0,0.252)(12.0,0.253)(13.0,0.253)(14.0,0.25)(15.0,0.252)(16.0,0.257)(17.0,0.261)(18.0,0.264)(19.0,0.266)(20.0,0.266)(21.0,0.267)(22.0,0.268)(23.0,0.268)};
                \addplot[
					color=red,
					mark=none,
					thick,
					densely dotted
					]
					coordinates { (1.0,0.166)(2.0,0.166)(3.0,0.165)(4.0,0.232)(5.0,0.348)(6.0,0.436)(7.0,0.507)(8.0,0.626)(9.0,0.667)(10.0,0.704)(11.0,0.724)(12.0,0.758)(13.0,0.784)(14.0,0.815)(15.0,0.862)(16.0,0.888)(17.0,0.888)(18.0,0.898)(19.0,0.904)(20.0,0.916)(21.0,0.924)(22.0,0.94)(23.0,0.993)};
				\addplot[
					color=black,
					mark=none,
					thick,
					densely dotted
					]
					coordinates { (1.0,0.19)(2.0,0.19)(3.0,0.189)(4.0,0.251)(5.0,0.352)(6.0,0.438)(7.0,0.507)(8.0,0.626)(9.0,0.667)(10.0,0.704)(11.0,0.724)(12.0,0.758)(13.0,0.784)(14.0,0.815)(15.0,0.862)(16.0,0.888)(17.0,0.888)(18.0,0.898)(19.0,0.904)(20.0,0.916)(21.0,0.924)(22.0,0.94)(23.0,0.993)};
                \end{axis}
            \end{tikzpicture}}
        \vspace{-0.25cm}
        \subfloat{
             \begin{tikzpicture}
            \pgfplotsset{every x tick label/.append style={font=\tiny}}
            \pgfplotsset{every y tick label/.append style={font=\tiny}}
                \begin{axis}[
                    width=0.55\linewidth,
                    height=4cm,
                    title={\footnotesize{n=5}},
                     title style={at={(0.063,0.915)},anchor=north,draw=black!60,fill=white},
                    ylabel={\scriptsize{$Acc(\lambda,LOG)$}},
                    xlabel near ticks,
                    ylabel near ticks,
                    xmin=0, xmax=24,
                    ymin=0, ymax=1.1,
                    xtick={0,1,2,3,4,5,6,7,8,9,10,11,12,13,14,15,16,17,18,19,20,21,22,23},
                    xticklabels={0,1,2,3,4,5,10,15,20,25,30,35,40,45,50,55,60,65,70,75,80,85,90,95},
                    ytick={0.2,0.4,0.6,0.8,1},
                    legend pos=north west,
                    ymajorgrids=false,
                    xmajorgrids=false,
                    major grid style={line width=.1pt,draw=gray!50},
                    x axis line style={draw=black!60},
                    tick style={draw=black!60},
                ]
                \addplot[
                    color=cyan,
	                mark=none,
	                thick,
	                dashed
	               ]
	                coordinates { (1.0,0.0)(2.0,0.0)(3.0,0.0)(4.0,0.077)(5.0,0.21)(6.0,0.303)(7.0,0.383)(8.0,0.537)(9.0,0.597)(10.0,0.657)(11.0,0.693)(12.0,0.74)(13.0,0.77)(14.0,0.793)(15.0,0.853)(16.0,0.883)(17.0,0.883)(18.0,0.897)(19.0,0.897)(20.0,0.897)(21.0,0.897)(22.0,0.917)(23.0,0.99)};
                \addplot[
					color=green!70,
					mark=none,
					thick,
					dashdotted
					]
					coordinates { (1.0,0.097)(2.0,0.097)(3.0,0.097)(4.0,0.136)(5.0,0.222)(6.0,0.276)(7.0,0.313)(8.0,0.327)(9.0,0.335)(10.0,0.343)(11.0,0.35)(12.0,0.351)(13.0,0.351)(14.0,0.349)(15.0,0.356)(16.0,0.369)(17.0,0.378)(18.0,0.385)(19.0,0.388)(20.0,0.388)(21.0,0.391)(22.0,0.391)(23.0,0.389)};
                \addplot[
					color=red,
					mark=none,
					thick,
					densely dotted
					]
					coordinates { (1.0,0.227)(2.0,0.227)(3.0,0.226)(4.0,0.288)(5.0,0.397)(6.0,0.491)(7.0,0.557)(8.0,0.66)(9.0,0.695)(10.0,0.722)(11.0,0.737)(12.0,0.765)(13.0,0.791)(14.0,0.824)(15.0,0.866)(16.0,0.89)(17.0,0.89)(18.0,0.899)(19.0,0.906)(20.0,0.923)(21.0,0.932)(22.0,0.947)(23.0,0.994)};
				\addplot[
					color=black,
					mark=none,
					thick,
					densely dotted
					]
					coordinates { (1.0,0.252)(2.0,0.252)(3.0,0.251)(4.0,0.307)(5.0,0.402)(6.0,0.492)(7.0,0.557)(8.0,0.66)(9.0,0.695)(10.0,0.722)(11.0,0.737)(12.0,0.765)(13.0,0.791)(14.0,0.824)(15.0,0.866)(16.0,0.89)(17.0,0.89)(18.0,0.899)(19.0,0.906)(20.0,0.923)(21.0,0.932)(22.0,0.947)(23.0,0.994)};
                \end{axis}
            \end{tikzpicture}}
        \subfloat{
             \begin{tikzpicture}
            \pgfplotsset{every x tick label/.append style={font=\tiny}}
            \pgfplotsset{every y tick label/.append style={font=\tiny}}
                \begin{axis}[
                    width=0.55\linewidth,
                    height=4cm,
                    title={\footnotesize{n=7}},
                    title style={at={(0.063,0.915)},anchor=north,draw=black!60,fill=white},
                    xlabel near ticks,
                    ylabel near ticks,
                    xmin=0, xmax=24,
                    ymin=0, ymax=1.1,
                    xtick={0,1,2,3,4,5,6,7,8,9,10,11,12,13,14,15,16,17,18,19,20,21,22,23},
                    xticklabels={0,1,2,3,4,5,10,15,20,25,30,35,40,45,50,55,60,65,70,75,80,85,90,95},
                    ytick={0.2,0.4,0.6,0.8,1},
                    legend pos=north west,
                    ymajorgrids=false,
                    xmajorgrids=false,
                    major grid style={line width=.1pt,draw=gray!50},
                    x axis line style={draw=black!60},
                    tick style={draw=black!60},
                ]
                \addplot[
                    color=cyan,
	                mark=none,
	                thick,
	                dashed
	               ]
	                coordinates { (1.0,0.0)(2.0,0.0)(3.0,0.0)(4.0,0.077)(5.0,0.21)(6.0,0.303)(7.0,0.383)(8.0,0.537)(9.0,0.597)(10.0,0.657)(11.0,0.693)(12.0,0.74)(13.0,0.77)(14.0,0.793)(15.0,0.853)(16.0,0.883)(17.0,0.883)(18.0,0.897)(19.0,0.897)(20.0,0.897)(21.0,0.897)(22.0,0.917)(23.0,0.99)};
                \addplot[
					color=green!70,
					mark=none,
					thick,
					dashdotted
					]
					coordinates { (1.0,0.097)(2.0,0.097)(3.0,0.097)(4.0,0.143)(5.0,0.249)(6.0,0.332)(7.0,0.38)(8.0,0.4)(9.0,0.412)(10.0,0.427)(11.0,0.441)(12.0,0.443)(13.0,0.446)(14.0,0.441)(15.0,0.447)(16.0,0.463)(17.0,0.477)(18.0,0.487)(19.0,0.491)(20.0,0.492)(21.0,0.494)(22.0,0.493)(23.0,0.494)};
                \addplot[
					color=red,
					mark=none,
					thick,
					densely dotted
					]
					coordinates { (1.0,0.27)(2.0,0.27)(3.0,0.27)(4.0,0.325)(5.0,0.427)(6.0,0.525)(7.0,0.587)(8.0,0.685)(9.0,0.714)(10.0,0.735)(11.0,0.746)(12.0,0.769)(13.0,0.789)(14.0,0.823)(15.0,0.866)(16.0,0.89)(17.0,0.891)(18.0,0.9)(19.0,0.909)(20.0,0.929)(21.0,0.942)(22.0,0.957)(23.0,0.995)};
				\addplot[
					color=black,
					mark=none,
					thick,
					densely dotted
					]
					coordinates { (1.0,0.286)(2.0,0.286)(3.0,0.285)(4.0,0.338)(5.0,0.431)(6.0,0.527)(7.0,0.587)(8.0,0.685)(9.0,0.714)(10.0,0.735)(11.0,0.746)(12.0,0.769)(13.0,0.789)(14.0,0.823)(15.0,0.866)(16.0,0.89)(17.0,0.891)(18.0,0.9)(19.0,0.909)(20.0,0.929)(21.0,0.942)(22.0,0.957)(23.0,0.995)};
                \end{axis}
            \end{tikzpicture}}
        \vspace{-0.25cm}
        \subfloat{
             \begin{tikzpicture}
            \pgfplotsset{every x tick label/.append style={font=\tiny}}
            \pgfplotsset{every y tick label/.append style={font=\tiny}}
                \begin{axis}[
                    width=0.55\linewidth,
                    height=4cm,
                    title={\footnotesize{n=9}},
                     title style={at={(0.063,0.915)},anchor=north,draw=black!60,fill=white},
                    xlabel={\scriptsize{$\lambda$ (\%)}},
                    ylabel={\scriptsize{$Acc(\lambda,LOG)$}},
                    xlabel near ticks,
                    ylabel near ticks,
                    xmin=0, xmax=24,
                    ymin=0, ymax=1.1,
                    xtick={0,1,2,3,4,5,6,7,8,9,10,11,12,13,14,15,16,17,18,19,20,21,22,23},
                    xticklabels={0,1,2,3,4,5,10,15,20,25,30,35,40,45,50,55,60,65,70,75,80,85,90,95},
                    ytick={0.2,0.4,0.6,0.8,1},
                    legend pos=north west,
                    ymajorgrids=false,
                    xmajorgrids=false,
                    major grid style={line width=.1pt,draw=gray!50},
                    x axis line style={draw=black!60},
                    tick style={draw=black!60},
                ]
                \addplot[
                    color=cyan,
	                mark=none,
	                thick,
	                dashed
	               ]
	                coordinates { (1.0,0.0)(2.0,0.0)(3.0,0.0)(4.0,0.077)(5.0,0.21)(6.0,0.303)(7.0,0.383)(8.0,0.537)(9.0,0.597)(10.0,0.657)(11.0,0.693)(12.0,0.74)(13.0,0.77)(14.0,0.793)(15.0,0.853)(16.0,0.883)(17.0,0.883)(18.0,0.897)(19.0,0.897)(20.0,0.897)(21.0,0.897)(22.0,0.917)(23.0,0.99)};
                \addplot[
					color=green!70,
					mark=none,
					thick,
					dashdotted
					]
					coordinates { (1.0,0.097)(2.0,0.097)(3.0,0.097)(4.0,0.143)(5.0,0.255)(6.0,0.381)(7.0,0.441)(8.0,0.467)(9.0,0.487)(10.0,0.503)(11.0,0.53)(12.0,0.53)(13.0,0.532)(14.0,0.515)(15.0,0.522)(16.0,0.543)(17.0,0.566)(18.0,0.582)(19.0,0.587)(20.0,0.588)(21.0,0.592)(22.0,0.595)(23.0,0.594)};
                \addplot[
					color=red,
					mark=none,
					thick,
					densely dotted
					]
					coordinates { (1.0,0.313)(2.0,0.313)(3.0,0.311)(4.0,0.366)(5.0,0.465)(6.0,0.568)(7.0,0.627)(8.0,0.711)(9.0,0.741)(10.0,0.753)(11.0,0.757)(12.0,0.776)(13.0,0.798)(14.0,0.834)(15.0,0.871)(16.0,0.893)(17.0,0.894)(18.0,0.901)(19.0,0.912)(20.0,0.939)(21.0,0.954)(22.0,0.966)(23.0,0.996)};
				\addplot[
					color=black,
					mark=none,
					thick,
					densely dotted
					]
					coordinates { (1.0,0.323)(2.0,0.323)(3.0,0.321)(4.0,0.374)(5.0,0.468)(6.0,0.569)(7.0,0.627)(8.0,0.711)(9.0,0.741)(10.0,0.753)(11.0,0.757)(12.0,0.776)(13.0,0.798)(14.0,0.834)(15.0,0.871)(16.0,0.893)(17.0,0.894)(18.0,0.901)(19.0,0.912)(20.0,0.939)(21.0,0.954)(22.0,0.966)(23.0,0.996)};
                \end{axis}
            \end{tikzpicture}}
        \subfloat{
             \begin{tikzpicture}
            \pgfplotsset{every x tick label/.append style={font=\tiny}}
            \pgfplotsset{every y tick label/.append style={font=\tiny}}
                \begin{axis}[
                    width=0.55\linewidth,
                    height=4cm,
                    title={\footnotesize{n=11}},
                    title style={at={(0.0745,0.915)},anchor=north,draw=black!60,fill=white},
                    xlabel={\scriptsize{$\lambda$ (\%)}},
                    xlabel near ticks,
                    ylabel near ticks,
                    xmin=0, xmax=24,
                    ymin=0, ymax=1.1,
                    xtick={0,1,2,3,4,5,6,7,8,9,10,11,12,13,14,15,16,17,18,19,20,21,22,23},
                    xticklabels={0,1,2,3,4,5,10,15,20,25,30,35,40,45,50,55,60,65,70,75,80,85,90,95},
                    ytick={0.2,0.4,0.6,0.8,1},
                    legend pos=north west,
                    ymajorgrids=false,
                    xmajorgrids=false,
                    major grid style={line width=.1pt,draw=gray!50},
                    x axis line style={draw=black!60},
                    tick style={draw=black!60},
                ]
                \addplot[
                    color=cyan,
	                mark=none,
	                thick,
	                dashed
	               ]
	                coordinates { (1.0,0.0)(2.0,0.0)(3.0,0.0)(4.0,0.077)(5.0,0.21)(6.0,0.303)(7.0,0.383)(8.0,0.537)(9.0,0.597)(10.0,0.657)(11.0,0.693)(12.0,0.74)(13.0,0.77)(14.0,0.793)(15.0,0.853)(16.0,0.883)(17.0,0.883)(18.0,0.897)(19.0,0.897)(20.0,0.897)(21.0,0.897)(22.0,0.917)(23.0,0.99)};
                \addplot[
					color=green!70,
					mark=none,
					thick,
					dashdotted
					]
					coordinates { (1.0,0.097)(2.0,0.097)(3.0,0.098)(4.0,0.152)(5.0,0.278)(6.0,0.398)(7.0,0.462)(8.0,0.494)(9.0,0.519)(10.0,0.54)(11.0,0.564)(12.0,0.564)(13.0,0.571)(14.0,0.55)(15.0,0.56)(16.0,0.577)(17.0,0.605)(18.0,0.624)(19.0,0.635)(20.0,0.637)(21.0,0.645)(22.0,0.647)(23.0,0.646)};
                \addplot[
					color=red,
					mark=none,
					thick,
					densely dotted
					]
					coordinates { (1.0,0.316)(2.0,0.316)(3.0,0.314)(4.0,0.366)(5.0,0.465)(6.0,0.565)(7.0,0.641)(8.0,0.736)(9.0,0.762)(10.0,0.773)(11.0,0.773)(12.0,0.783)(13.0,0.797)(14.0,0.833)(15.0,0.872)(16.0,0.893)(17.0,0.894)(18.0,0.902)(19.0,0.915)(20.0,0.947)(21.0,0.965)(22.0,0.977)(23.0,0.997)};
				\addplot[
					color=black,
					mark=none,
					thick,
					densely dotted
					]
					coordinates { (1.0,0.322)(2.0,0.322)(3.0,0.32)(4.0,0.37)(5.0,0.467)(6.0,0.565)(7.0,0.641)(8.0,0.736)(9.0,0.762)(10.0,0.773)(11.0,0.773)(12.0,0.783)(13.0,0.797)(14.0,0.833)(15.0,0.872)(16.0,0.893)(17.0,0.894)(18.0,0.902)(19.0,0.915)(20.0,0.947)(21.0,0.965)(22.0,0.977)(23.0,0.997)};
                \end{axis}
            \end{tikzpicture}}
            \vspace{-0.1cm}
    \ref{plotLabel5}\vspace{-0.3cm}
    \caption{Mean accuracy of the Goal Mirroring (GM), Naive Bayes Model (NBM) approaches and the two hybrid approaches Weighted Sum (WS) and Tiebreaking (TB) on the \emph{logistics domain} for different sizes of the training set $n$.}
    \label{fig:ComparisonArtificialDomains}
\end{figure*}
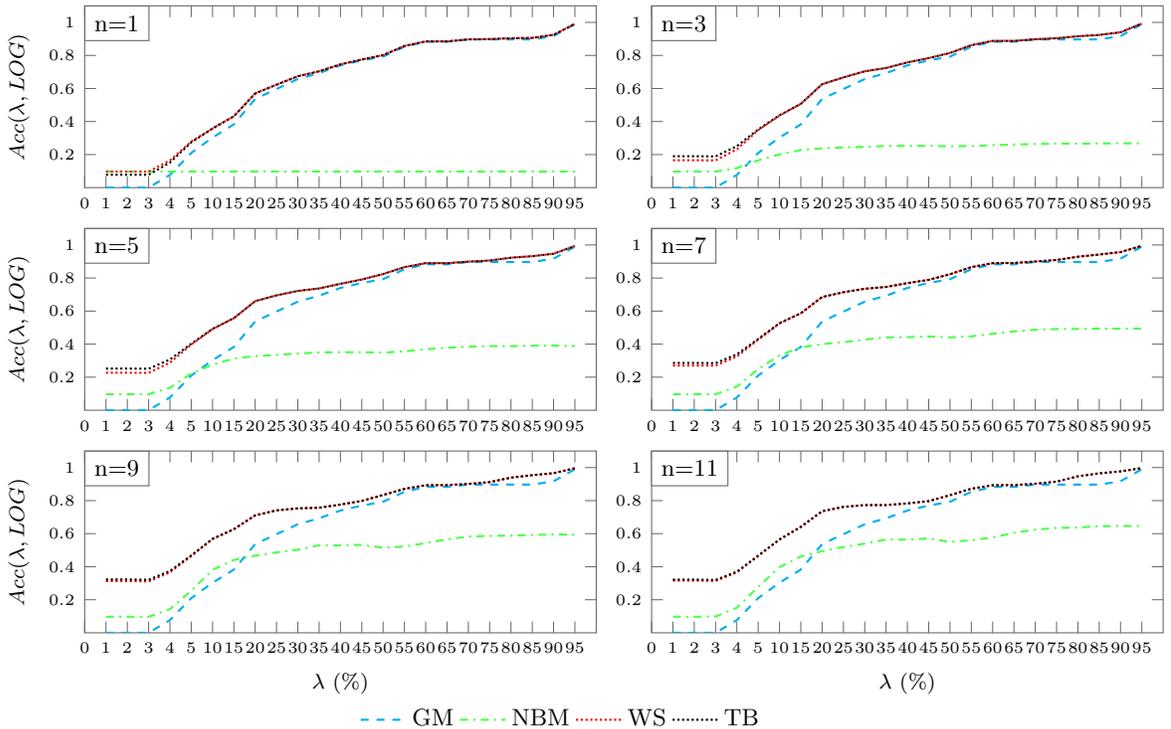